\documentclass{acmart}
\usepackage{url}
\usepackage{float}
\usepackage{pgfplots}
\pgfplotsset{compat=1.18}
\usepackage{tikz}
\usetikzlibrary{positioning, arrows.meta, calc}
\usepackage{xcolor}

\usepackage{graphicx}
\usepackage{todonotes}
\usepackage[british]{babel}
\usepackage{csquotes}
\usepackage{subcaption}
\MakeOuterQuote{"}
\usepackage{pdfpages}
\usepackage{colortbl}
\usepackage{tcolorbox}

\title{Beyond Generation: Using Multimodal LLMs as Detectors of LLM-Generated Deepfake Facial Images}

\author{Olivia Salamon}
\author{Virginia N. L. Franqueira}
\email{V.Franqueira@kent.ac.uk}
\orcid{0000-0003-1332-9115}
\author{Shujun Li}
\email{S.J.Li@kent.ac.uk}
\orcid{0000-0001-5628-7328}

\affiliation{%
  \\
  \institution{Institute of Cyber Security for Society (iCSS) \& School of Computing, University of Kent, London and South East University Group}
  \country{Canterbury, UK}
  }
       
\begin{document}  

\begin{abstract}
\textbf{Background/Aim}: As image generation models become increasingly capable of producing highly realistic content, traditional forensic techniques are struggling to keep up. This paper investigates whether MLLMs can instead be used as a form of forensic analysis by identifying synthetic media. Specifically, the study aims to determine whether these models detect fake images by recognising physical inconsistencies, or whether they rely on digital artefacts such as embedded watermarks.

\noindent \textbf{Methods}: Six MLLMs: GPT-4o, GPT-5, GPT-5.2, Gemini 2.5, Gemini 3, and Seed 1.6, are benchmarked using a dataset of 400 images. Fifty images are authentic photographs, while the remaining 350 are generated using seven leading image models: Gemini 3, Gemini 2.5, Flux.2 Pro, GPT-Image-1, GPT-Image-1.5, Seedream 4.0, and Seedream 4.5. The evaluation is structured in three stages: single-image classification, pairwise comparison, and a focused ``self-detection'' experiment on Gemini. This final stage tests five image groups, including clean controls, visible logos, and images containing invisible SynthID watermarks, to explore how digital signals influence detection accuracy. 

\noindent \textbf{Results}: Several MLLMs achieved a high detection accuracy, with GPT-5.2 reaching a peak of 97.9\% in Experiment 2, the pairwise experiment, against specific generators and an overall mean pairwise accuracy of 82.5\%  by identifying structural or physical inconsistencies within images. However, results from the self-detection experiment indicate that the presence of invisible SynthID watermarks leads to a 10\% reduction in Gemini's detection accuracy compared to the clean control group. One possible explanation is that SynthID embedding reduces some of the visible artefacts the models rely on for detection.

\noindent \textbf{Conclusion}: These results imply that MLLMs can detect structural artefacts without relying on watermarks, but performance varies substantially across model generations. The findings from comparison C5 also suggest that SynthID embedding may suppress some of the structural artefacts used during detection.
\end{abstract}



\maketitle  

\section{Introduction} %
\label{sec:introduction}

\subsection{Background and Motivation}

Generative tools such as OpenAI’s GPT-series and Google's Gemini-series have become increasingly advanced in their image generation capabilities. Throughout this study, the term deepfake refers to fully synthetic facial images generated from text prompts using the seven generative models: Gemini 3, Gemini 2.5, Flux.2 Pro, GPT-Image-1, GPT-Image-1.5, Seedream 4.0, and Seedream 4.5. This study focuses specifically on fully synthetic deepfakes, images generated entirely from text prompts, rather than face-swap or identity-manipulation techniques, which involve modifying existing footage. While generative models are increasingly capable of both approaches, this study isolates text-to-image synthesis as the primary forensic threat. These models now produce images that are difficult to distinguish from real photographs. Traditional forensic methods are becoming inadequate for this task; methods that once relied on pixel-level inconsistencies~\cite{rossler2019} are becoming obsolete. Researchers have increasingly turned to MLLMs for deepfake detection, through their abilities to “justify” their detection classifications. Rather than searching for pixel artefacts, research has shown that MLLMs can identify semantic inconsistencies within images such as irregular lighting or inconsistent background geometry~\cite{jia2024}. This study evaluates whether recent models such as GPT-5.2 demonstrate similar or improved capability.

While MLLM decision justification represents a promising step for Explainable AI (XAI), these models remain unpredictable in their classification behaviour. Lapuschkin et al.~\cite{lapuschkin2019} and Maheronnaghsh~\cite{maheronnaghsh2025} highlight the “Clever Hans” effect, when a model achieves high accuracy by exploiting shortcuts within the image rather than a genuine forensic understanding. These shortcuts often stem from the generator's traces such as visible logos or pixel-embedded watermarks such as Google’s SynthID.

This study assesses the reliability of MLLMs as forensic detectors. By isolating and removing generator-related traces within Experiment 3, and comparing detection performance across conditions in Experiments 1 and 2, this study examines the extent to which MLLM accuracy may depend on structural image artefacts or digital shortcuts.

Central to this research is the audit of MLLM decision-making. It remains unclear whether MLLMs genuinely reason about image structure or whether high accuracy reflects exploitation of digital shortcuts, such as metadata cues or pixel-embedded watermarks, rather than genuine forensic analysis~\cite{maheronnaghsh2025,ji2025}. Current evidence is not yet sufficient to answer this question, motivating the controlled experimental design used in this study.

\subsection{Research Hypotheses}
To achieve this, the following four Hypotheses have been established:

\begin{itemize}
    \item \textbf{H1 (Self- and Cross-Model Detection):} Generative ecosystems may have an advantage in detecting their own generated images, however, detection performance against images from external models is expected to vary significantly across architectures~\cite{ji2025}.
    \item \textbf{H2 (Pairwise Image Advantage):} Detection accuracy is theoretically higher in a pairwise experiment, where the MLLM is presented with one real and one fake image simultaneously, than in a single-image experiment~\cite{jia2024,chandra2025}.
    \item \textbf{H3 (Effects of Watermarks):} The presence or absence of watermark signals will have a measurable effect on detection accuracy~\cite{liu2024,han2025}.
\item \textbf{H4 (Justification Reliability):} MLLM natural-language justifications may not accurately reflect the reasoning process, raising concerns about whether MLLM justifications are genuine structural analysis~\cite{maheronnaghsh2025}.
\end{itemize}

\subsection{Research Objectives}
These hypotheses are further reinforced within the following objectives evaluated throughout the study:
\begin{itemize}
    \item \textbf{Dataset Construction:} Building a custom 400-image corpus, 50 authentic photographs from the Flickr-Faces-HQ dataset~\cite{karras2019} in conjunction with 350 images from seven top-ranked generators, including the Gemini Ecosystem, GPT family, Flux and SeeDream.
    \item \textbf{Detector Performance:} Evaluating six MLLM detectors across single-image and pairwise classification experiments.
    \item \textbf{Gemini Watermark Experiments:} Performing an “ablation” experiment on the Gemini generated images by removing their visible logos and SynthID embedded watermarks to measure how this affects detection accuracy.
    \item \textbf{Model Justification Audit:} Manually validating the natural-language justifications provided by the model to determine if their justifications are reasonable or if they are “hallucinating” artefacts within the image.
\end{itemize}

\subsection{Report Structure}

\begin{itemize}
    \item \textbf{Section 2: Literature Review} - Identifies the current gaps discovered around how MLLMs justify their decisions.
    \item \textbf{Section 3: Methodology} - Details the technical framework established for the study. This includes the development of the image dataset used and the experiment designs for the single-image analysis, pairwise-image analysis and ablation experiments.
    \item \textbf{Section 4: Results} - Presents the core findings of the study and comparative performance of the chosen MLLM detectors.
    \item \textbf{Section 5: Discussion} - Discusses the key findings, positions them within the literature, and addresses limitations and future directions.
    \item \textbf{Section 6: Conclusion} - Final closing statement of the study.
\end{itemize}

\section{Literature Review}
\label{sec:literature_review}
Generative AI ecosystems such as Gemini and the GPT family can now produce media that is genuinely difficult to detect. Successful detection now requires techniques such as the use of MLLMs as opposed to conventional pixel-based algorithms.

\subsection{Evolution of Generative AI}
Within the early days of generative media, it relied heavily on deep learning architectures such as Generative Adversarial Networks (GANs)~\cite{goodfellow2014generative} which produced media that was easily detectable both through the human eye and AI. The introduction of pre-trained language models such as BERT~\cite{devlin2019} set the foundation for advanced language models. This was then further developed through Latent Diffusion Models (LDMs)~\cite{rombach2022} which produced high-resolution images through a denoising technique and Large Vision Models (LVMs), which extended this to visual understanding. This has led to the formation of Multimodal Large Language Models (MLLMs) which are capable of processing both text and image inputs.

\subsection{Historical Overview of Deepfakes}
Deepfakes have evolved through several generations, the first being video-based face-swap techniques~\cite{rossler2019} which produced visible artefacts detectable by pixel-level analysis. The second generation improved realism through GAN-based models to generate synthetic facial images. The third generation, introduced by~\cite{rombach2022}, generated high-fidelity images from textual prompts with a significant decline in detectable inconsistencies, enabled by LDMs. The most recent generation involves MLLMs generating images from multimodal prompts, further reducing image detectability~\cite{ricker2024towards}. This study focuses on the third and fourth generations of synthetic media however, deepfake videos remain a significant concern. 

\subsection{Automated Deepfake Detection}
As generative quality has significantly improved, detection strategies have shifted to high-level semantic evaluation. MLLMs have increasingly been applied to zero-shot deepfake classification. Jia et al.~\cite{jia2024} demonstrated that models within the GPT and Gemini families could identify artificial media by identifying inconsistencies within the image, rather than depending on pixel-level artefacts. These models may detect issues such as inconsistent lighting or unrealistic object structures.

Although the idea of using MLLMs as deepfake detectors initially seemed promising, newer studies have produced contradictory results. Large-scale testing has revealed that although they show some baseline detection capability, models experience a large drop in accuracy when tested against high-quality, real, diffusion images~\cite{ren2025,chandra2025}. In addition, Jia et al.~\cite{jia2024} argued that the current MLLMs are not yet reliable enough for forensic detection, demonstrating that even a slight image alteration can drastically alter their judgement. These inconsistent findings suggest that the models tested were likely looking for visual shortcuts rather than genuine forensic reasoning.

Specifically, no recent study has used a controlled ablation design to isolate whether the accuracy of MLLM detection reflects genuine structural reasoning or the exploitation of watermarks, which this study directly addresses within Experiment 3.

\subsection{Watermarks and Their Importance}
Due to the increased challenges created by modern generative models, digital watermarks are increasingly promoted as a practical detection solution. Traditional image metadata can be easily manipulated by the user. Therefore, image watermarking techniques work through embedding signals directly into the image architecture~\cite{evsutin2022}. 

Google DeepMind's SynthID is a leading example of this technology. Gowal et al.\cite{gowal2025} outlined how the tool embeds an invisible digital signature directly into the image's frequency domain, theoretically overcoming the traditional trade-off between visual quality and signal strength ensuring the watermark persists even through common image transformations such as cropping or compression. According to DeepMind's developers, this ensured that even if an image is altered, the watermark remained detectable~\cite{dathathri2024scalable}.

However, recent security audits have shown that embedded signals are far more fragile than the industry seems to claim. Liu et al.~\cite{liu2024} found that invisible watermarks, such as systems similar to SynthID, could still be targeted by attackers through a ``controllable regeneration'' attack. This method removes hidden patterns without compromising the quality of the image. Han et al.~\cite{han2025} explained this weakness, finding that even basic image alterations can cause SynthID signals to deteriorate. Relying on embedded watermarks alone therefore creates a vulnerability to the Trace Removal Attack (TRA) described by Liu et al.\cite{liuC2022}, enabling attackers to generate seemingly real fakes that bypass modern detection systems.

\subsection{Explainability and Justification Quality}
MLLM usage has noticeably increased due to their Explainable AI (XAI) capabilities, providing natural language justifications alongside their classifications~\cite{li2025fakescope}. Knowing why a classification was made is arguably as important as the classification itself, especially in forensic scenarios~\cite{qian2025blackboxes}. 

However, analysing these explanations exposes a major flaw: the ``Clever Hans'' effect. Maheronnaghsh and Alvanagh~\cite{maheronnaghsh2025} argued that MLLMs often achieve high accuracy scores not because of their capabilities to recognise facial anomalies but instead ``cheat'' through finding visual shortcuts within the testing data to find the correct answer. Ji et al.~\cite{ji2025} demonstrated self-detection bias, explained that while these models had higher success in recognising fakes built through their own architecture, they struggle against previously unseen diffusion models and rely on irrelevant visual traces to make a guess. 

Beyond the models taking shortcuts, a second problem has emerged: LLM hallucinations. When asked to justify a decision, they can often fabricate supporting evidence. Both Sun et al.~\cite{sun2025} and Li et al.~\cite{li2025} demonstrated that models will confidently output factually incorrect descriptions. An example seen is the model flagging a non-existent distortion that simply is not evident within the image. Rather than the model performing true analysis, it has actually constructed a believable but fabricated justification. To extensively test the model's reasoning capabilities, a Two-Alternative Forced Choice (TAFC) framework~\cite{green1966} has been adopted within this study.

\subsection{Ethics, Diversity and Industry Context} 
How a dataset is constructed directly affects what a detection model learns to look for~\cite{Zhangetal2024}. Dataset bias is a relevant problem within digital forensics as it limits the models' ability to evaluate diverse demographic groups accurately~\cite{Linetal2024}. Widely available datasets such as FFHQ~\cite{karras2019} remain the industry standard for training Generative AI; researchers must also acknowledge that FFHQ does not accurately represent certain age groups or ethnic backgrounds with approximately 70\% of images featuring Caucasian individuals~\cite{luccioni2023}. Bias towards certain age brackets, ethnic backgrounds, and environmental lighting directly leads to higher false-positive rates for under-represented groups. 

The synthetic corpus was designed using a prompt matrix to ensure demographic variance across age, gender, ethnicity and professional job roles to reduce the possibility of stereotypical outputs. Similarly, the FFHQ control group images were manually selected to ensure the evaluation framework did not favour any particular demographic profile. Together, the diverse prompt matrix and the selection of FFHQ images were implemented to avoid demographic shortcuts that can make MLLMs unreliable forensic tools.

\section{Methodology and Implementation}
 \label{sec:methodology}
This section outlines the methodology and implementation choices applied across three experiments. Each experiment is designed to address one or more of the four hypotheses established in Section~\ref{sec:introduction}. A concern across all experiments is the distinction between self-model detection, where a detector evaluates images generated by its own ecosystem, and cross-model detection, where a detector evaluates images from an external generator. This self-/cross-model comparison is integrated into experiments 1 and 2 through the full 6×7 detector-generator matrix, and is the primary focus of H1.

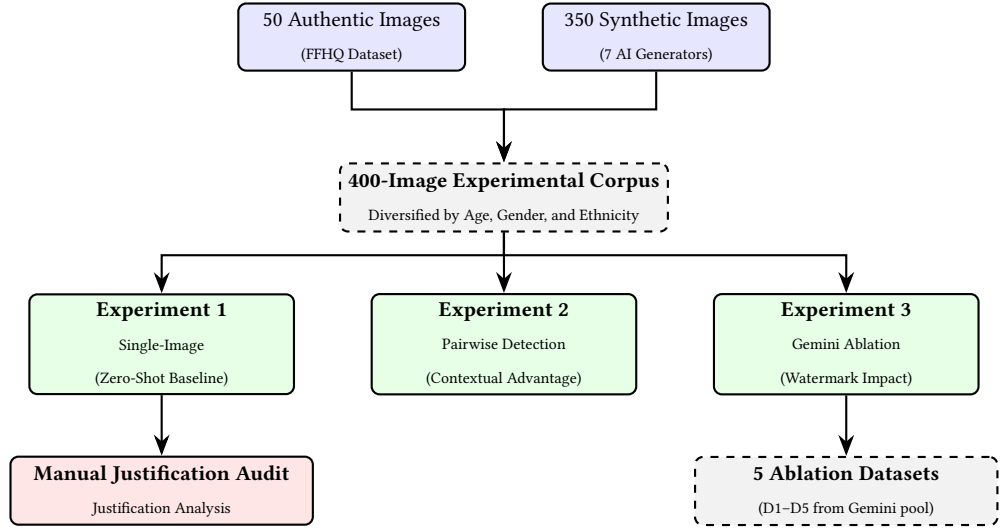
\begin{figure}[H]
\centering
\begin{tikzpicture}[
    node distance=0.8cm and 1cm,
    base/.style={draw, thick, align=center, rounded corners=3pt, font=\small},
    source/.style={base, fill=blue!10, minimum width=3cm},
    pool/.style={base, fill=gray!10, minimum width=4cm, dashed},
    exp/.style={base, fill=green!10, minimum width=3.5cm, minimum height=1cm},
    audit/.style={base, fill=red!10, minimum width=4cm},
    arrow/.style={-{Stealth[scale=1.2]}, thick}
]

    \node [source] (real) {50 Authentic Images \\ \scriptsize (FFHQ Dataset)};
    \node [source, right=of real] (fake) {350 Synthetic Images \\ \scriptsize (7 AI Generators)};

    \node [pool, below=1.2cm of $(real.south)!0.5!(fake.south)$] (corpus) {\textbf{400-Image Experimental Corpus} \\ \scriptsize Diversified by Age, Gender, and Ethnicity};

    \node [exp, below=of corpus] (e2) {\textbf{Experiment 2} \\ \scriptsize Pairwise Detection \\ \scriptsize (Contextual Advantage)};
    \node [exp, left=of e2] (e1) {\textbf{Experiment 1} \\ \scriptsize Single-Image \\ \scriptsize (Zero-Shot Baseline)};
    \node [exp, right=of e2] (e3) {\textbf{Experiment 3} \\ \scriptsize Gemini Ablation \\ \scriptsize (Watermark Impact)};
    \node [pool, below=of e3] (d1d5) {\textbf{5 Ablation Datasets} \\ \scriptsize (D1--D5 from Gemini pool)};
    \draw [arrow] (e3.south) -- (d1d5.north);

    \node [audit, below=of e1] (hitl) {\textbf{Manual Justification Audit} \\ \scriptsize Justification Analysis};

    \draw [arrow] (real.south) -- ++(0,-0.5) -| (corpus.north);
    \draw [arrow] (fake.south) -- ++(0,-0.5) -| (corpus.north);
    
    \draw [arrow] (corpus.south) -- ++(0,-0.3) -| (e1.north);
    \draw [arrow] (corpus.south) -- (e2.north);
    \draw [arrow] (corpus.south) -- ++(0,-0.3) -| (e3.north);
    
    \draw [arrow] (e1.south) -- (hitl.north);

\end{tikzpicture}
\caption{Methodological Framework: The logical progression from data sourcing (FFHQ and AI Generators) through the three-tiered experimental structure, concluding with the qualitative audit for Experiment 1 justifications.}
 \Description{}
\label{fig:method_overview}
\end{figure}

The structure of this methodology is illustrated below within Figure~\ref{fig:method_overview}.

\subsection{Dataset Construction}

All materials, including the datasets, used in this study are available at: \url{https://github.com/OliviaSalamon/dissertation-corpus}.

Prior testing for this study revealed a significant problem to overcome: images generated by older models, such as Gemini 1.5, contained inconsistencies that would result in the MLLMs classifying them as AI-generated, such as overly smooth skin textures or unusually perfectly glossed image backgrounds~\cite{rombach2022}.

The dataset was constructed in three stages. First, a synthetic corpus of 350 images was generated across seven leading generators, each generator providing a pool of 50 images, using a structured prompt matrix. Second, the prompt matrix was designed using a diversity-guided Equality, Diversity, and Inclusion (EDI) framework to prevent stereotypical outputs. Third, a manually selected control group of 50 authentic photographs was sourced from FFHQ~\cite{karras2019}.

\subsubsection{Synthetic Corpus Generation}
To overcome this limitation, a custom 400-image dataset was constructed, 350 of which were generated using seven state-of-the-art text-to-image generators. The models selected for this study were ranked among the top performers on the Artificial Analysis Text to Image Leaderboard~\cite{artificialanalysis2025} at the time of dataset creation (December 2025). The seven generators used within this study were: Gemini 3, Gemini 2.5, Flux.2 Pro, GPT-Image-1, GPT-Image-1.5, Seedream 4.0, and Seedream 4.5.

All 350 images were generated via each generator's official developer API, the generation scripts are provided in the project corpus 

(\texttt{01\_Pre\_Experimental\_Setup/generate\_images\_master\_script.py}). 
Generating these images through APIs ensured a baseline without UI-specific watermarks, user-interface alterations, or image compression that can result from web interface generation. The only exception to this API-only pipeline occurs in Experiment 3, where additional images were specifically generated using the Gemini Web Interface to capture the visible logo watermark for comparative ablation testing.

\subsubsection{Prompt Matrix Design}
A structured prompt matrix (full prompt matrix is available in the project corpus at 
(\texttt{01\_Pre\_Experimental\_Setup/project\_prompts.csv}), was constructed to ensure tests across the different models remain consistent. This was established to prevent stereotypical outputs that would directly impact detection rates of the images~\cite{luccioni2023}. To mitigate this, a dynamic prompt structure was used to ensure contextual variance within variables such as age, gender, physical traits, ethnicity, and professional roles were embedded into each of the 50 prompts within the prompt matrix. For example, prompts included: `A professional headshot of a woman in her 50s with grey hair, natural window light, centred composition' and `A teenage boy with South Asian features, casual outdoor setting, overcast lighting.' 
In addition to contextual variance, photograph background bias was further mitigated by applying consistent photographic qualifiers, such as ``natural window light'' and ``centred composition''. The prompt matrix was designed to prevent any demographic shortcuts the generator models might rely on. 

\subsubsection{Real Image Control Group} 
A control group of 50 authentic images of human faces was sourced from the Flickr-Faces-HQ (FFHQ) dataset~\cite{karras2019}. FFHQ was selected for a few main reasons: its images are taken under real-world conditions at high resolution (1024×1024) making them a representative and authentic baseline. In addition, its portrait composition is directly comparable to AI-generated head shots. However, the base FFHQ dataset is known to be heavily demographically biased, consisting of nearly 70\% Caucasian and only 4.2\% Black individuals~\cite{luccioni2023}. To mitigate this, images were manually selected to guarantee a balance of age groups, genders and ethnicities. This purposeful sampling prevented the AI detectors from using demographic features as a classification shortcut.

\subsection{Experimental framework} 
To evaluate the forensic capabilities of the chosen MLLMs, a three-tiered framework was developed with each experiment designed to address specific hypotheses. Experiment 1 (single-image zero-shot classification) provides the baseline detection data required to evaluate H1 (Self- and Cross-Model Detection) and H4 (Justification Reliability). Experiment 2, pairwise TAFC classification, is designed to test H2 (Pairwise Image Advantage) by comparing accuracy against the Experiment 1 baseline. Experiment 3 (Gemini watermark ablation) directly addresses H3 (Effects of Watermarks) by isolating the contribution of visible and invisible watermark signals to detection accuracy.

\subsubsection{Experiment 1: Single-Image Classification} 
\paragraph{Phase A: Quantitative Zero-Shot Baseline} 
The first experiment was designed to establish a zero-shot baseline. In this experiment, each MLLM was presented with a single image, randomly drawn from either the 50 authentic FFHQ images or from one of seven generator-specific pools, each containing 50 images produced by a single generator. 
\paragraph{Phase B: Qualitative Forensic Reasoning} 
Building directly onto the single-image classification within Experiment 1 and to move testing beyond a binary result (1 for correct classification, 0 for incorrect classification) this phase of Experiment 1 focuses on an automation script to investigate the “black box” of AI decision-making~\cite{qian2025blackboxes}. Phase B extracts the natural-language justifications provided by each MLLM, forming the evidence base for the H4 (Justification Reliability) evaluation. The validity of these justifications is assessed in Phase C. This consisted of a subset of 10 fixed images (5 real images from the FFHQ dataset and 5 AI-generated images).
This subset size was selected to allow five independent classification runs per image per model, producing 50 data points per model for consistent analysis while remaining feasible for the Manual Justification Audit process outlined below in Phase C. The five AI-generated images were drawn from four distinct generator sources: Flux.2 Pro, GPT-Image-1.5, Gemini 3, and GPT-Image-1, to ensure the reasoning evaluation was carried across the top modern generation architectures. By presenting the same ten images to each model across five independent runs, this phase tested whether each model produced consistent reasoning for identical inputs; inconsistent justifications across runs would highlight inaccuracies within detection rather than genuine structural analysis. As the MLLMs outputted their reasoning in a structured JSON format, the script allowed for analysis to check whether visible or invisible watermark signals were referenced in the model's text outputs. 

\paragraph{Phase C: Manual Justification Audit} 
To conclude the single-image analysis experiment, the final phase involved a Manual Justification Audit of the natural-language justifications produced in Phase B. Since MLLMs are known to `hallucinate' plausible but factually incorrect justifications~\cite{ji2025,bang2023}, the outputs from Phase B were manually audited to evaluate whether the reasoning provided was accurate. 
Raw justifications were cross-referenced against a manual visual audit. The MLLM responses were categorised as `Correct + Reasonable' if they referenced correct cues, or `Hallucination' if the classification was correct but supported by non-existent or incorrect artefacts. Manually validating the justifications ensured the study's conclusions reflected genuine forensic evaluations by the MLLM rather than a statistical guess. 

\subsubsection{Experiment 2: Pairwise Image Analysis (Contextual Advantage)} 
The second experiment introduced contextual reasoning through a TAFC evaluation protocol~\cite{green1966}. To ensure the results were not affected by positional bias, the placement of the authentic and synthetic images as either ``Image A'' or ``Image B'' was fully randomised.

In this part of the experiment, the MLLMs were provided with two randomly selected images, one real image from the FFHQ control dataset and one synthetic image generated by one of the 7 generators at random from the 7 pools of 50 images. The models were prompted with ``You are presented with two images, labelled IMAGE A and IMAGE B. One is a REAL photograph, and the other is AI-GENERATED. Your task is to identify which image is the AI-generated fake'', in order to identify which of the two images was AI-generated and which was a real photograph, providing a binary classification and a justification for their decision.

This methodology, which focuses on the comparison was directly designed to test the Contextual Advantage Hypothesis (H2). The experiment measured whether accuracy improved when models were provided with a real reference image alongside the synthetic one, testing whether comparative context benefits detection performance. Justification analysis was not conducted for Experiment 2, as reasoning auditing was limited to Phase B and C of Experiment 1.

\subsubsection{Experiment 3: Targeted Watermark Ablation Study (Gemini/SynthID)} 
\paragraph{Phase A: Forensic Rationale and Dataset Construction} 
By introducing an ablation approach, cropping visible watermarks and resampling invisible ones through pixel resampling, this experiment investigates the true detection capabilities of MLLMs. The objective is to determine whether these models genuinely recognise deepfake artefacts, or if their accuracy is influenced by generator traces; a methodology that aligns with the framework established by Gandhi and Jain~\cite{gandhi2020}.

The five-dataset comparison framework within this study isolates each signal: visible logo, invisible SynthID, and their combination, to determine whether MLLM accuracy reflects genuine forensic reasoning or watermark exploitation.
This decision to perform an ablation study on the Gemini ecosystem was driven by the integration of SynthID, Google DeepMind’s invisible Watermarking tool. According to official documentation, SynthID ``embeds a digital watermark directly into the pixels of an image, making it imperceptible to the human eye, but detectable for identification''~\cite{deepmind2023}.

Preliminary research for this study revealed a double layer forensic system within Gemini: images generated via the Gemini Web Interface not only include the invisible SynthID embedded watermark but also a visible watermark on the lower right hand corner of the generated image, a small, four-pointed ``sparkle'' icon. Images sourced directly from the Gemini API (Imagen 3) do not generate the visible logo yet only with SynthID embedded within the image. This provided an opportunity to study the importance of each marker in a controlled environment, forming the basis for the five-dataset ablation structure.

To investigate this, five distinct datasets (Pools) were constructed:
\begin{itemize}
\item \textbf{D1 (Gemini-API):} Raw Gemini API output (Imagen 3) — SynthID only, no logo. 
\textit{These images form the Gemini generator pool used in Experiments 1 and 2.}
\item \textbf{D2 (Gemini-Web):} Raw Gemini Web Interface output — SynthID + visible sparkle logo.
\item \textbf{D3 (Gemini-Web-Cropped):} D2 with visible logo manually cropped.
\item \textbf{D4 (Gemini-Web-Ablated):} D2 subjected to LDM reconstruction — logo remains, SynthID disrupted.
\item \textbf{D5 (Gemini-Full-Ablation):} D2 cropped and subjected to LDM reconstruction — no logo, SynthID disrupted.
\end{itemize}

\paragraph{Phase B: Dataset Structure and Ablation Design} 
To complete the ablation experiment, the neutralisation of the SynthID signal in Datasets 4 and 5 was essential. This was achieved through an Image-to-Image (img2img) LDM reconstruction pass utilising the Hugging Face \textit{Diffusers} library~\cite{vonPlaten2022diffusers}. Specifically, the pipeline employed the Realistic Vision V5.1 model~\cite{davi2023realistic}, which was selected for its ability to maintain high photographic quality during pixel resynthesis.

By altering the pixels at a tested 0.15 strength, a value supported by Zhao et al.~\cite{zhao2023invisible} for diffusion-based image alterations, the high-frequency digital watermarks were successfully disrupted while still maintaining high visual quality of the image's original state. This specific strength was selected to minimise degradation of the subject’s facial biometrics and keeping the image metadata such as the background and lighting as similar as possible to the originally generated image. 
This could be a tactic a malicious actor might use to bypass AI detection by running an image through a secondary generative pass. The complete watermark removal script is provided in the corpus at \texttt{04\_Experiment\_3\_Watermark\_Ablation/remove\_synthid\_dataset4.py}.

\paragraph{Phase C: Comparison Matrix}
To evaluate the impact of specific forensic markers (H3), five distinct dataset comparisons were conducted. This matrix isolates the presence of invisible SynthID signals against visible ``sparkle'' watermarks to determine their individual and combined effects on detection accuracy. The raw dataset comparison results are archived within the corpus at 
\texttt{04\_Experiment\_3\_Watermark\_Ablation/Dataset\_Comparison\_Results/}.

\begin{itemize}
    \item \textbf{C1 (API vs. Cropped Web):} D1 (Gemini-API) vs. D3 (Gemini-Web-Cropped). Evaluates consistency across API and Web-generated sources when visible markers are absent.
    \item \textbf{C2 (Visible Logo Effect):} D2 (Gemini-Web) vs. D3 (Gemini-Web-Cropped). Measures the accuracy gain provided by the visible logo when the invisible signal is present.
    \item \textbf{C3 (Logo-Only Detection):} D4 (Gemini-Web-Ablated) vs. D5 (Gemini-Full-Ablation). Assesses if the detector relies on the visible logo once the invisible signal is neutralised.
    \item \textbf{C4 (Invisible Signal Effect):} D2 (Gemini-Web) vs. D4 (Gemini-Web-Ablated). Isolates the effectiveness of the SynthID watermark while the visible logo remains constant.
    \item \textbf{C5 (Invisible-Only vs. Control):} D3 (Gemini-Web-Cropped) vs. D5 (Gemini-Full-Ablation). Evaluates detection performance based solely on the invisible signal against a fully ablated control.
\end{itemize}

\begin{figure}[H]
     \centering
     \begin{subfigure}[t]{0.19\textwidth}
         \centering
         \includegraphics[width=\textwidth]{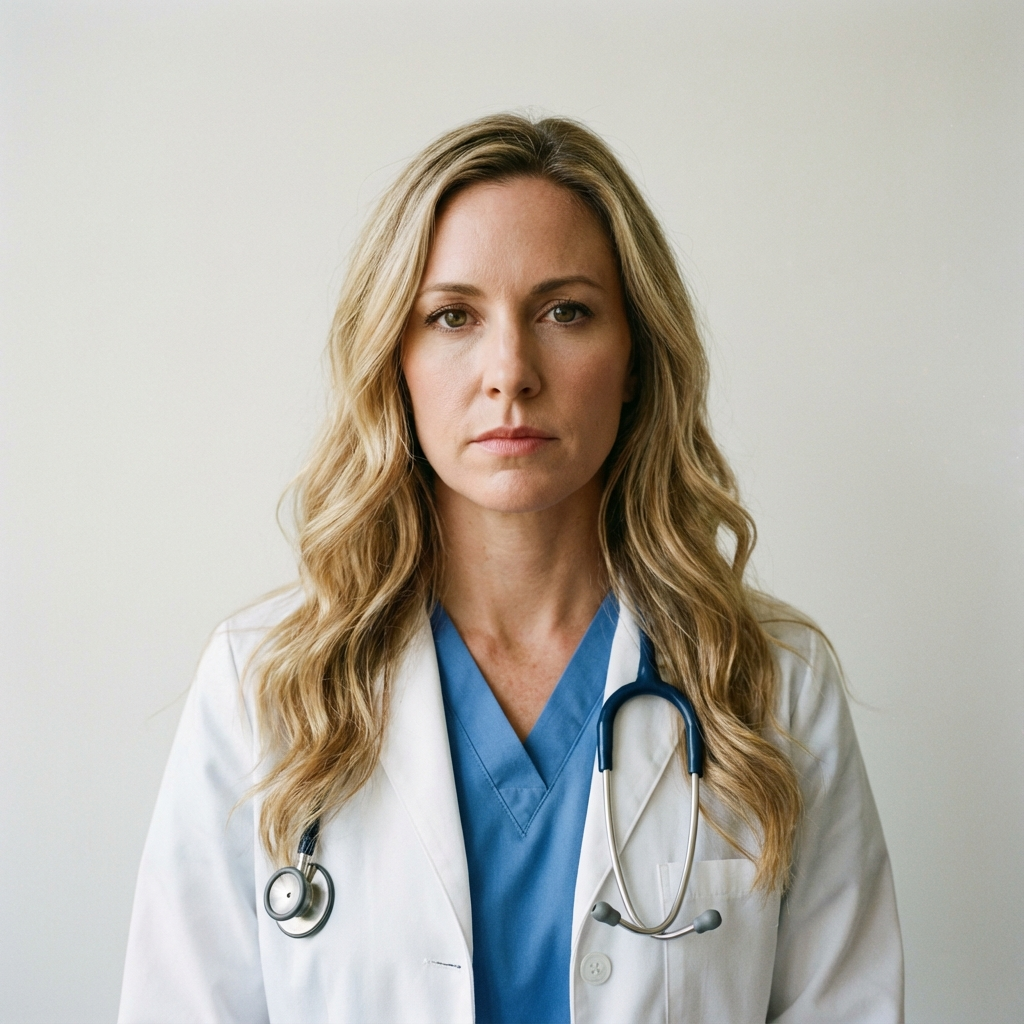}
         \caption*{\scriptsize \textbf{D1:} SynthID Only}
     \end{subfigure}
     \hfil
     \begin{subfigure}[t]{0.19\textwidth}
         \centering
         \includegraphics[width=\textwidth]{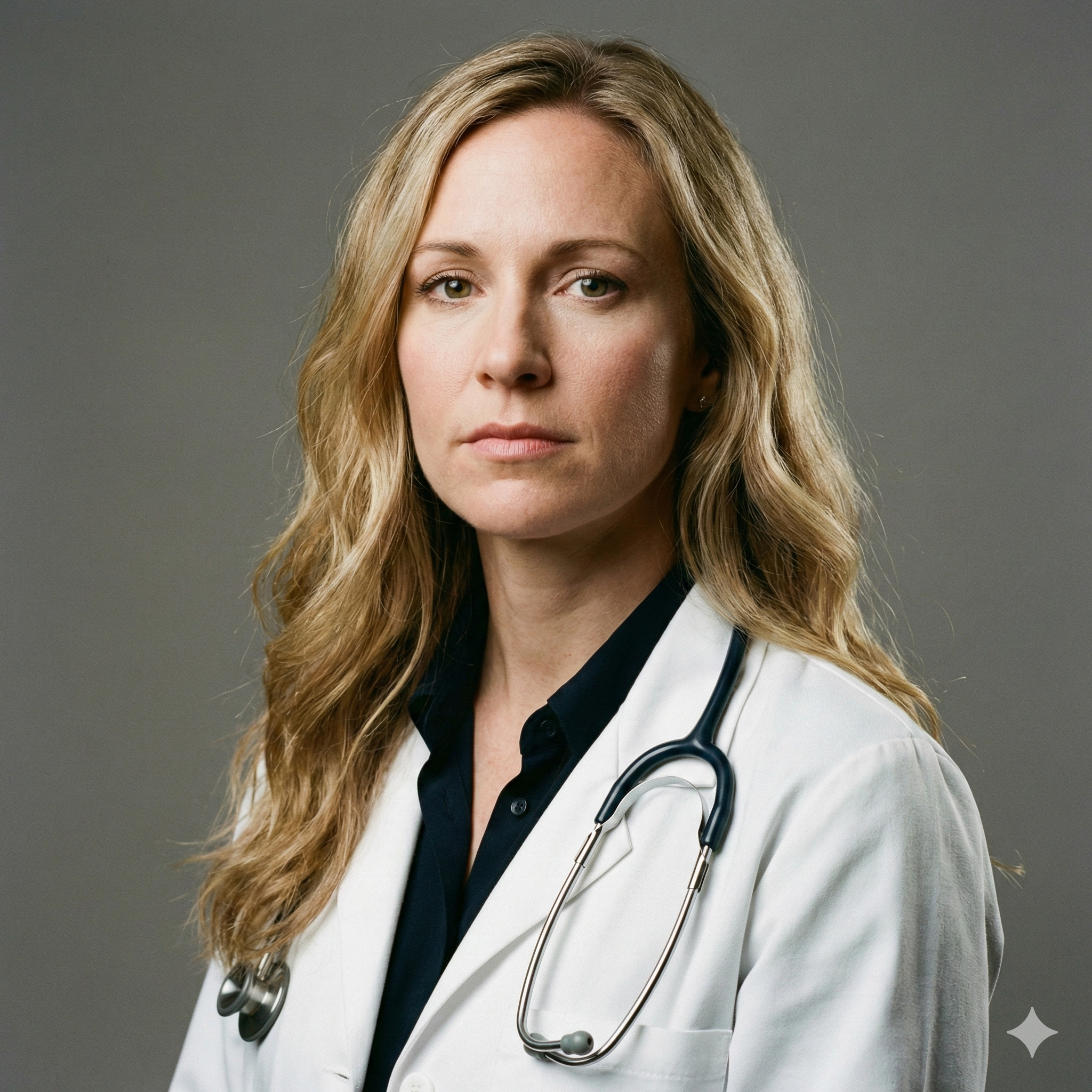}
         \caption*{\scriptsize \textbf{D2:} Logo + SynthID}
     \end{subfigure}
     \hfil
     \begin{subfigure}[t]{0.19\textwidth}
         \centering
         \includegraphics[width=\textwidth]{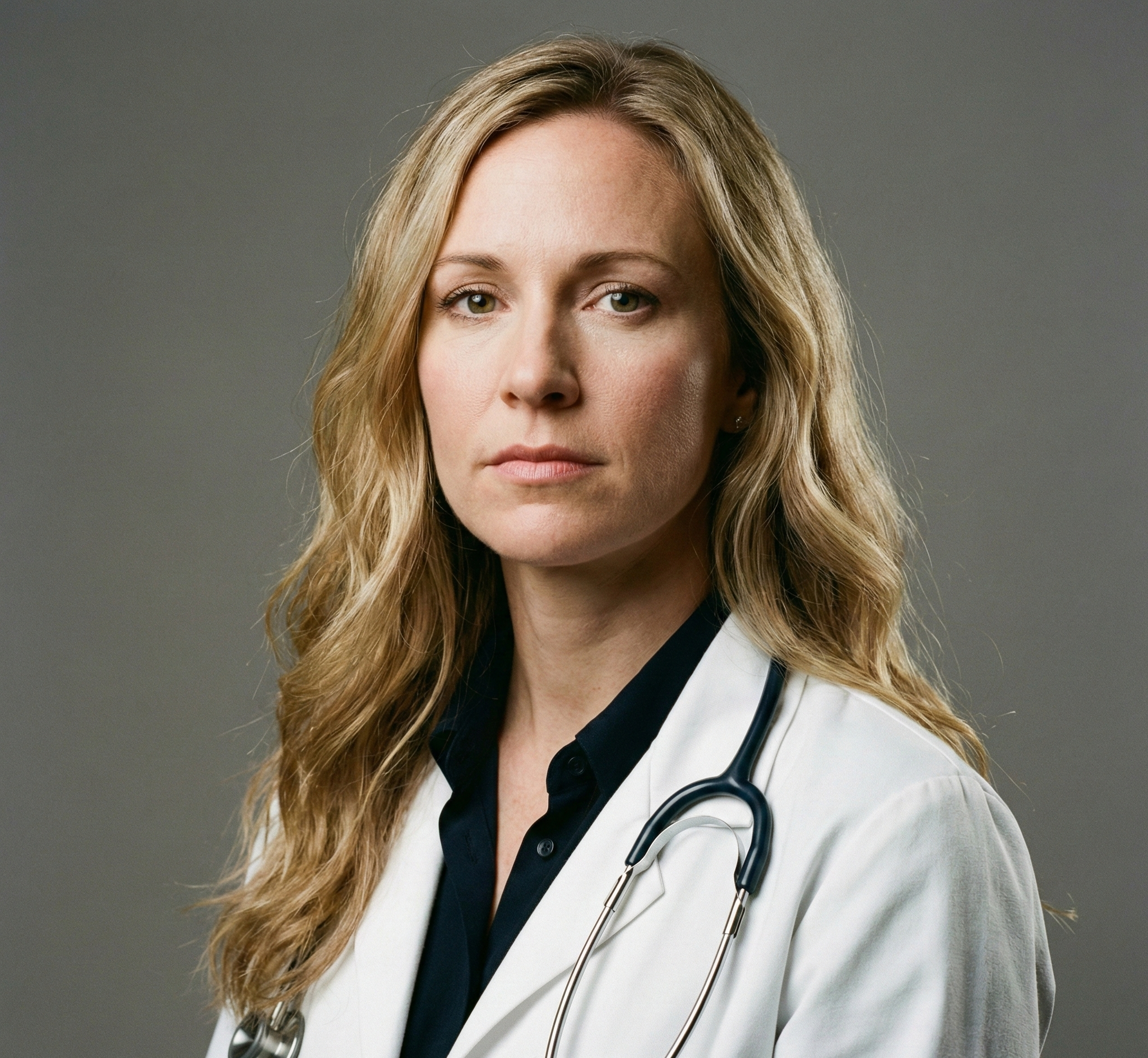}
         \caption*{\scriptsize \textbf{D3:} Logo Removed}
     \end{subfigure}
     \hfil
     \begin{subfigure}[t]{0.19\textwidth}
         \centering
         \includegraphics[width=\textwidth]{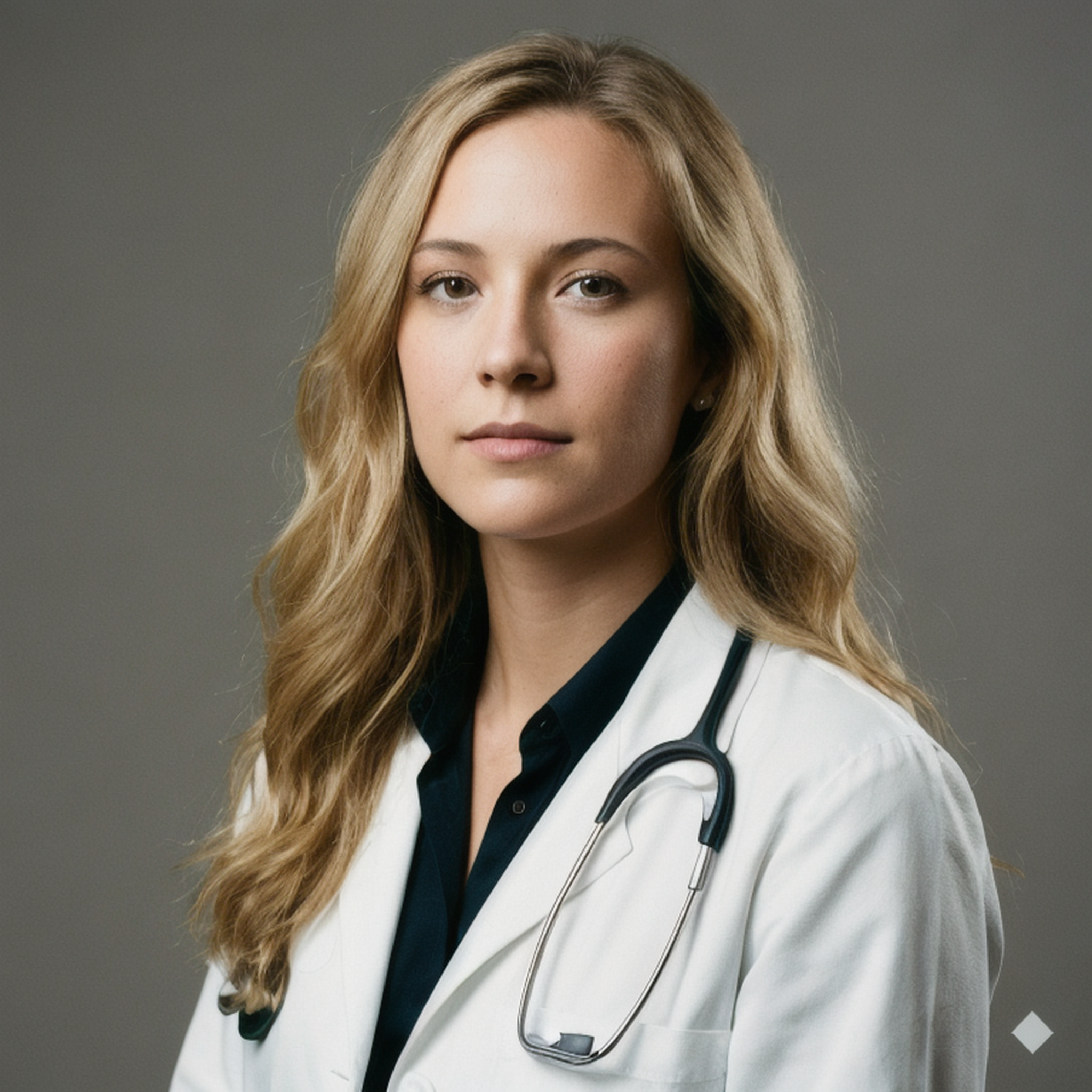}
         \caption*{\scriptsize \textbf{D4:} SynthID Disrupted}
     \end{subfigure}
     \hfil
     \begin{subfigure}[t]{0.19\textwidth}
         \centering
         \includegraphics[width=\textwidth]{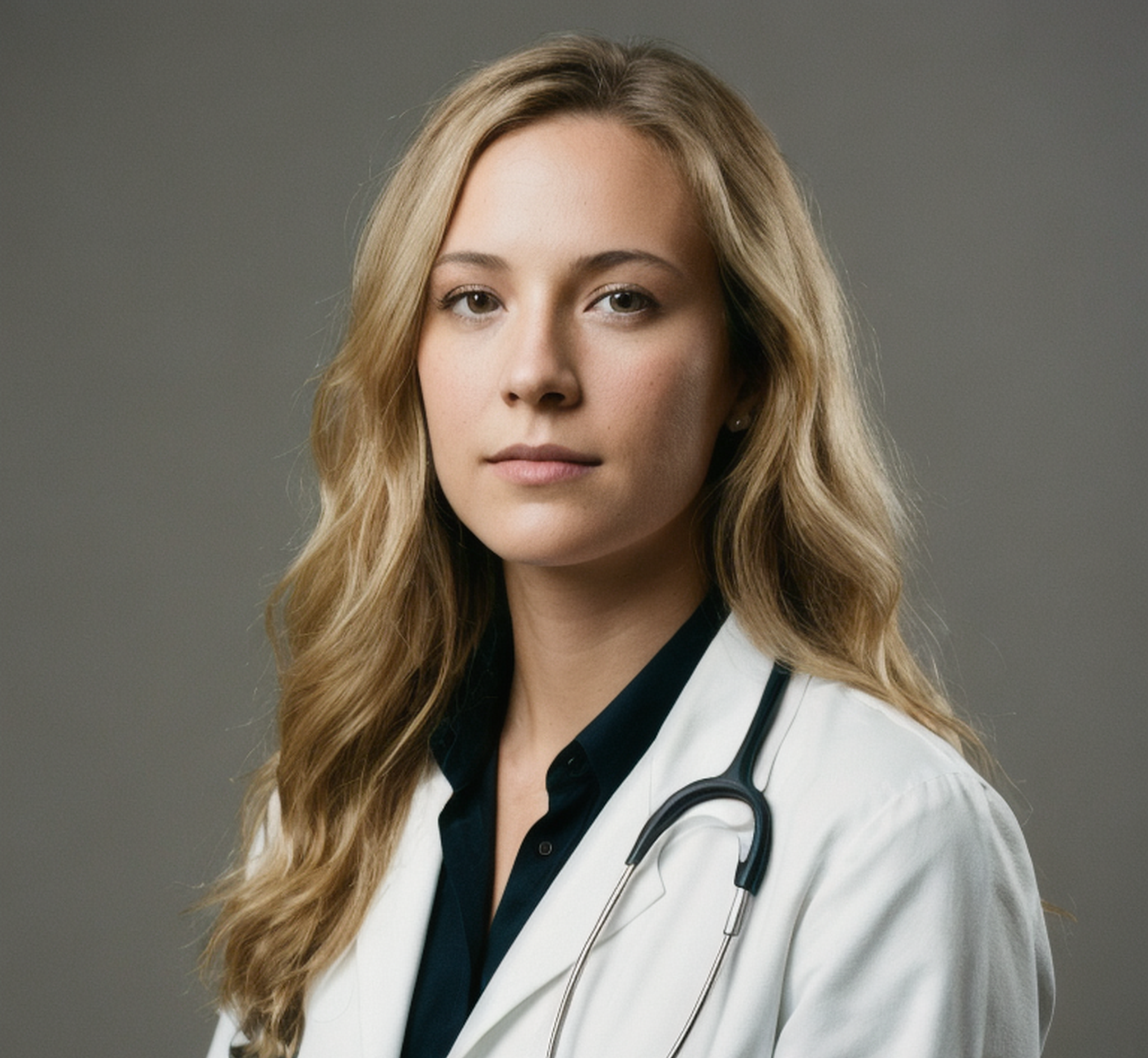}
         \caption*{\scriptsize \textbf{D5:} Full-ablation}
     \end{subfigure}
     
    \caption{Targeted Watermark Ablation Series: Visual comparison of the five Gemini dataset pools (D1–D5) used to isolate the forensic impact of visible logos and invisible digital signatures.}
     \Description{}
    \label{fig:dataset_comparison}
\end{figure}

Figure~\ref{fig:dataset_comparison} below provides a visual comparison of the five datasets within the ablation study.

\subsection{Implementation Framework and Parameter Design} 

\subsubsection{Automation Pipeline}
The experimental phases were executed through a custom Python pipeline interacting with the OpenAI~\cite{openai_api}, Google Gemini~\cite{google_api}, and Replicate~\cite{replicate_api} APIs. The single-image, pairwise, and justification evaluation scripts are provided in the corpus at 

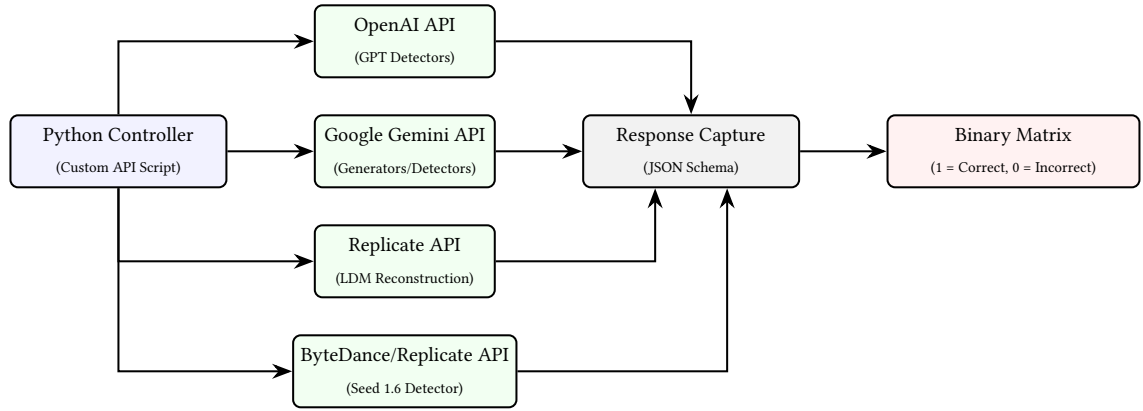
\begin{figure}[H]
\centering
\resizebox{\textwidth}{!}{
\begin{tikzpicture}[
    node distance=1.2cm and 1.2cm, 
    base/.style={draw, thick, align=center, rounded corners=3pt, font=\small},
    proc/.style={base, fill=blue!5, minimum width=3cm, minimum height=1cm},
    api/.style={base, fill=green!5, minimum width=2.5cm, minimum height=1cm},
    data/.style={base, fill=gray!10, minimum width=3cm, minimum height=1cm},
    result/.style={base, fill=red!5, minimum width=3.5cm, minimum height=1cm},
    arrow/.style={-{Stealth[scale=1.2]}, thick}
]

    \node [proc] (controller) {Python Controller \\ \scriptsize (Custom API Script)};
    
    \node [api, right=of controller] (google) {Google Gemini API \\ \scriptsize (Generators/Detectors)};
    \node [api, above=0.5cm of google] (openai) {OpenAI API \\ \scriptsize (GPT Detectors)};
    \node [api, below=0.5cm of google] (replicate) {Replicate API \\ \scriptsize (LDM Reconstruction)};
    \node [api, below=0.5cm of replicate] (seed) {ByteDance/Replicate API \\ \scriptsize (Seed 1.6 Detector)};
    
    \node [data, right=of google] (json) {Response Capture \\ \scriptsize (JSON Schema)};
    
    \node [result, right=of json] (matrix) {Binary Matrix \\ \scriptsize (1 = Correct, 0 = Incorrect)};

    \draw [arrow] (controller.north) |- (openai.west);
    \draw [arrow] (controller.east) -- (google.west);
    \draw [arrow] (controller.south) |- (replicate.west);
    \draw [arrow] (controller.south) |- (seed.west);
    
    \draw [arrow] (openai.east) -| (json.north);
    \draw [arrow] (google.east) -- (json.west);
    
    \draw [arrow] (replicate.east) -| ([xshift=-0.5cm]json.south);
    \draw [arrow] (seed.east) -| ([xshift=0.5cm]json.south);
    
    \draw [arrow] (json.east) -- (matrix.west);

\end{tikzpicture}
} 
\caption{Automation Pipeline: Workflow illustrating the interaction between the custom Python controller, the four external API services (including the Seed 1.6 detector), and the automated data extraction process.}
 \Description{}
\label{fig:automation_flow}
\end{figure}

\texttt{02\_Experiment\_1\_Single\_Image/} and \texttt{03\_Experiment\_2\_Pairwise/}.
Replicate was specifically utilised to host the LDM reconstruction models required for the watermark ablation study in Experiment 3. Findings were automatically extracted via JSON schemas into a binary matrix (1 for correct, 0 for incorrect) for statistical analysis.
The architectural logic of this automated framework and the data extraction process are visualized in Figure~\ref{fig:automation_flow}.

\subsubsection{Parameter Standardisation and Bias Control}
Temperature settings were applied differently across generators and detectors. For image generation, a temperature of 0.0 was set where the API permitted within an adjustable parameter to produce deterministic outputs. For the six MLLM detectors, models were kept at their default settings (typically 1.0) to represent realistic conditions. The temperature differences between models were a deliberate choice to allow detectors to operate under their standard conditions. Not all MLLM APIs expose a temperature parameter; where this setting was unavailable, the model's default configuration was used. Gemini's detection configuration was explicitly set to 0.0 to ensure consistent justification outputs across the five repeated runs in Phase B of Experiment 1.
Instructions and system prompts remained uniform across all platforms to prevent any potential prompt bias.

For Experiment 2, the pipeline utilised a shuffle randomisation algorithm~\cite{habibzadeh2015shuffle} to randomise the positioning of images (Position A vs. Position B). This neutralised positional bias inherent in TAFC protocols~\cite{macmillan2004, Hautus2021}.

\subsection{Statistical Methodology} 

The independent variables across the three experiments were:

\begin{itemize}
    \item \textbf{Detector Model:} The six MLLMs under evaluation, including GPT-5.2, GPT-5, GPT-4o, Gemini 3, Gemini 2.5, and Seed 1.6.
    \item \textbf{Generator Source:} the generator source of the evaluated image (the seven generators plus FFHQ authentic images)
    \item \textbf{Experimental Condition:} the experimental condition: single-image, pairwise, or ablation dataset tier (D1–D5).
\end{itemize}
The dependent variable in all three experiments was binary classification accuracy: 1 for a correct classification and 0 for an incorrect one. 
Since binary classification scores (0 or 1) follow a Bernoulli distribution and do not meet the normality assumptions required for parametric testing, a non-parametric framework was adopted to evaluate the hypotheses at a significance level of $p < 0.05$:

\begin{itemize}
    \item \textbf{Mann-Whitney U Test:} Applied to independent samples to compare baseline detection rates across distinct model families (e.g., OpenAI vs. Google).
    \item \textbf{Wilcoxon Signed-Rank Test:} Applied to dependent, paired samples to measure how an individual model's performance changed under specific experimental variables, such as pairwise presentation or watermark ablation. The Python scripts implementing these tests are provided at \texttt{04\_Experiment\_3\_Watermark\_Ablation/Gemini\_Statistics.py}, with raw outputs at 
    
    \texttt{04\_Experiment\_3\_Watermark\_Ablation/wilcoxon\_manwhitneyu\_results.txt}.
\end{itemize}

Finally, to ensure that any statistically significant findings held practical weight in an experimental context, the Rank-Biserial Correlation ($r$) was used to calculate the effect size.

\section{Results}
\label{sec:results}

\subsection{Overview of Experimental Results}
Section~\ref{sec:methodology} described image generators as Large Vision Models (LVMs) while the six MLLM detectors: GPT-5.2, GPT-5, GPT-4o, Gemini 3, Gemini 2.5 and Seed 1.6 represent a separate set of models. During the time of evaluation, GPT-4o and Seed 1.6 were included as detectors because they represent the most accessible commercial MLLMs from their ecosystems. This enabled cross-generation comparison within the GPT family and cross-ecosystem comparison with ByteDance's Seed platform.

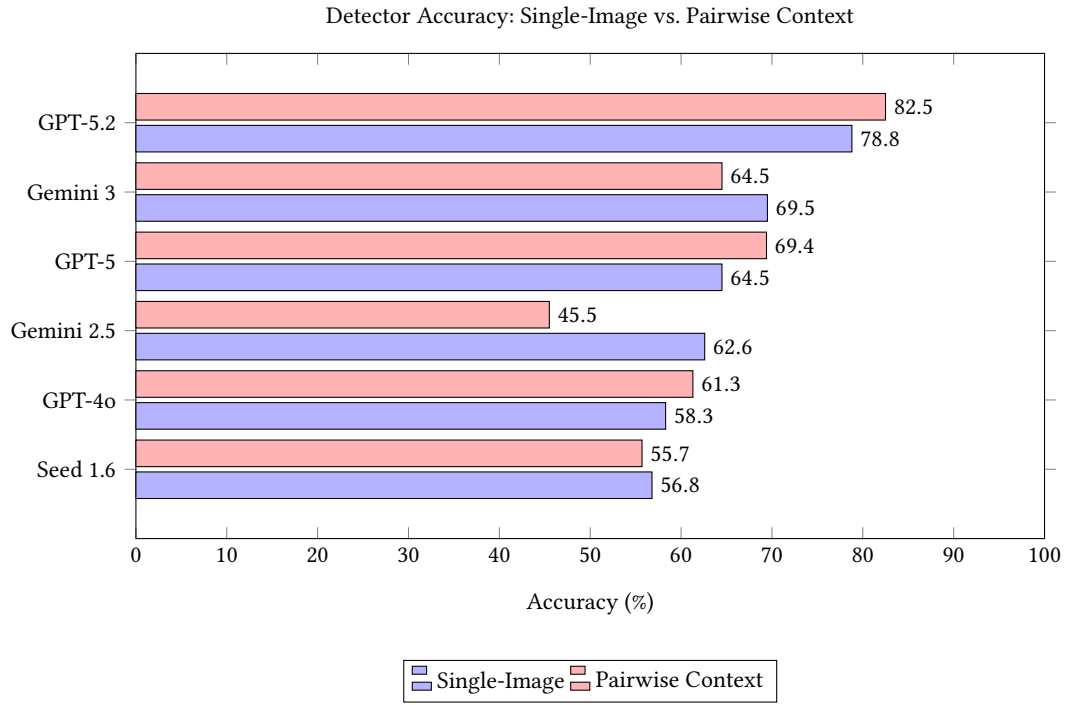
\begin{figure}[ht]
\centering
\begin{tikzpicture}
\begin{axis}[
    xbar,
    width=0.9\textwidth,
    height=8cm,
    xmin=0, xmax=100,
    xlabel={Accuracy (\%)},
    xlabel style={yshift=-5pt}, 
    symbolic y coords={Seed 1.6, GPT-4o, Gemini 2.5, GPT-5, Gemini 3, GPT-5.2}, 
    ytick=data,
    nodes near coords,
    nodes near coords align={horizontal},
    legend style={at={(0.5,-0.25)}, anchor=north, legend columns=-1},
    title={Detector Accuracy: Single-Image vs. Pairwise Context},
    bar width=10pt,
    enlarge y limits=0.2,
]
\addplot[fill=blue!30] coordinates {
    (78.8,GPT-5.2) (69.5,Gemini 3) (64.5,GPT-5) 
    (62.6,Gemini 2.5) (58.3,GPT-4o) (56.8,Seed 1.6)
};
\addplot[fill=red!30] coordinates {
    (82.5,GPT-5.2) (64.5,Gemini 3) (69.4,GPT-5) 
    (45.5,Gemini 2.5) (61.3,GPT-4o) (55.7,Seed 1.6)
};
\legend{Single-Image, Pairwise Context}
\end{axis}
\end{tikzpicture}
\caption{Comparative performance of detector models. Results are ranked by baseline single-image mean from best to worst. Note the performance gains in the GPT family compared to the accuracy decline observed in the Gemini ecosystem under pairwise conditions.}
 \Description{}
\label{fig:performance_chart}
\end{figure}

Table~\ref{tab:overview} provides a cross-experiment overview of pairwise versus single-image accuracy shifts across all six detectors, directly relevant to H2 (Pairwise Image Advantage). Figure~\ref{fig:performance_chart} visualises the single-image versus pairwise accuracy gap across all six detectors.

\begin{table}[H] 
\centering
\caption{Summary of detector model performance across single-image and pairwise conditions. Models are ranked in descending order according to their mean single-image accuracy.}
\label{tab:overview}
\resizebox{\textwidth}{!}{ 
\begin{tabular}{lcccc}
\hline
\textbf{Model} & \textbf{Single Mean (\%)} & \textbf{Pairwise Mean (\%)} & \textbf{Context Impact} & \textbf{SD (Single)} \\
\hline
GPT-5.2    & 78.8 & 82.5 & +3.7 & 3.41 \\
Gemini 3   & 69.5 & 64.5 & -5.0 & 4.67 \\
GPT-5      & 64.5 & 69.4 & +4.9 & 5.23 \\
Gemini 2.5 & 62.6 & 45.5 & -17.1 & 4.59 \\
GPT-4o     & 58.3 & 61.3 & +3.0 & 4.46 \\
Seed 1.6   & 56.8 & 55.7 & -1.1 & 6.70 \\
\hline
\end{tabular}
} 
\end{table}


\subsection{Experiment 1 Results: Single-Image Baseline and Qualitative Reasoning}
\subsubsection{Quantitative Zero-Shot Baseline}
Tier 1’s primary objective was to determine whether MLLMs are capable of identifying AI-generated images without any external references. The performance matrix within Table~\ref{tab:single_matrix} highlights that some models demonstrated clear forensic capability; the remaining models struggled to exceed even the 50\% baseline for this study. 
Based on the single-image analysis results, the most effective detector was GPT-5.2 achieving a mean accuracy of 78.8\%. GPT-5 and Gemini 3 also performed well with levels above 60\% (full data within the corpus \texttt{02\_Experiment\_1\_Single\_Image/single\_image\_analysis\_phaseA.xlsx}).

\begin{table}[ht]
\centering
\caption{Single-image detection accuracy (\%) per detector model per generator source. Models are grouped by vendor ecosystem. Shaded cells indicate same-ecosystem detector-generator pairs (self-detection), relevant to H1.}
\label{tab:single_matrix}
\resizebox{\textwidth}{!}{ 
\begin{tabular}{lccccccc}
\hline
\textbf{Detector} & \textbf{Flux.2 Pro} & \textbf{GPT-Img-1} & \textbf{GPT-Img-1.5} & \textbf{Gem-2.5} & \textbf{Gem-3} & \textbf{SD-4.0} & \textbf{SD-4.5} \\
\hline
GPT-5.2    & 87.9 & \cellcolor{yellow!25}84.3 & \cellcolor{yellow!25}72.1 & 84.6 & 65.8 & 78.6 & 78.6 \\
GPT-5      & 88.6 & \cellcolor{yellow!25}51.4 & \cellcolor{yellow!25}50.7 & 91.9 & 45.0 & 52.1 & 71.4 \\
GPT-4o     & 78.6 & \cellcolor{yellow!25}60.7 & \cellcolor{yellow!25}37.9 & 76.5 & 36.2 & 56.4 & 61.4 \\
\hline
Gemini 3   & 72.1 & 74.3 & 72.1 & \cellcolor{yellow!25}67.1 & \cellcolor{yellow!25}67.8 & 68.6 & 64.3 \\
Gemini 2.5 & 82.9 & 80.0 & 44.3 & \cellcolor{yellow!25}79.9 & \cellcolor{yellow!25}40.3 & 54.3 & 56.4 \\
\hline
Seed 1.6   & 73.6 & 45.7 & 50.0 & 76.5 & 47.7 & 47.1 & 57.1 \\
\hline
\textbf{Mean} & 80.6 & 66.1 & 54.5 & 79.4 & 50.4 & 59.5 & 64.9 \\
\hline
\end{tabular}
}
\end{table}
However, the results revealed a significant performance gap with both Seed 1.6 (56.8\%) and GPT-4o (58.3\%) hovering just above the 50\% baseline used. These models often failed to identify structural artefacts within the images,  suggesting that their detection capabilities are not precise enough to be used within a forensic setting. In addition the Standard Deviation (SD) for Seed 1.6 (6.70) was the highest figure amongst the models. This indicated that not only was the model's performance low but also inconsistent compared to the other models tested across the ten experimental runs. Therefore, while H1 is supported for the more advanced architectures, detection capability is not consistent across the current MLLMs.
The results further revealed that detection accuracy was significantly influenced by the generator source of the AI-generated image. As shown in Table~\ref{tab:single_matrix}, the detection rates varied considerably depending on which of the seven generators produced the image that was being evaluated. These results suggest that generator architecture has influence on how detectable the image is.

With a cross-model mean detection rate of 80.6\%, Flux.2 Pro was the most easily identifiable generator. The results imply that Flux.2 Pro likely produces more detectable artefacts.
In contrast, Gemini 3 generated images were the most difficult to detect by the six detectors within the study, with a mean detection rate of just 50.4\%, at the boundary of random chance and therefore indicating that the most recent generative architectures are reaching a point where standalone zero-shot single-image classification is becoming an unreliable forensic methodology.

A Mann-Whitney U test comparing self-model detection accuracy (same-ecosystem detector-generator pairs) against cross-model accuracy revealed $p = 0.384$, $r = 0.188$, confirming that ecosystem alignment was not a statistically significant predictor of detection accuracy.

The diagonal effect, where same-ecosystem detector-generator pairs show varying performance, is visible in the highlighted cells of Table~\ref{tab:single_matrix} and is discussed in the H1 verdict.

\subsubsection{Qualitative Reasoning}
Phase B investigated the forensic reasoning provided by each model through an evaluation focused on a subset of ten images, each analysed across five separate iterations. Using a Manual Justification Audit, justifications were sorted into specific categories: “Correct + Reasonable” were assigned when a model identified authentic structural indicators, while “Incorrect Detection” was used for failed classifications. The compiled results are presented in Table~\ref{tab:hitl}; the complete audit spreadsheet is provided in the corpus at \texttt{02\_Experiment\_1\_Single\_Image/Manual\_audit.xlsx}.

\begin{table}[H] 
\centering
\caption{Phase B Manual Justification Audit results: frequency of correct and reasonable justifications across 48 evaluated cases per model (10 fixed images × 5 runs = 50 classification events per model; 2 cases excluded due to Gemini 3's content safety block on one authentic image, yielding 48 usable responses) Models are presented in a fixed order consistent with baseline performance rankings.}
\label{tab:hitl}
\resizebox{0.8\textwidth}{!}{ 
\begin{tabular}{lccc}
\hline
\textbf{Model} & \textbf{Correct + Reasonable} & \textbf{Incorrect Detection} & \textbf{Accuracy (\%)} \\
\hline
GPT-5.2    & 45 & 3  & 93.8 \\
Gemini 3   & 37 & 11 & 77.1 \\
GPT-5      & 31 & 17 & 64.6 \\
Gemini 2.5 & 37 & 11 & 77.1 \\
GPT-4o     & 28 & 20 & 58.3 \\
Seed 1.6   & 29 & 19 & 60.4 \\
\hline
\end{tabular}
} 
\end{table} 

\begin{figure}[h]
    \centering
    \includegraphics[width=0.3\linewidth]{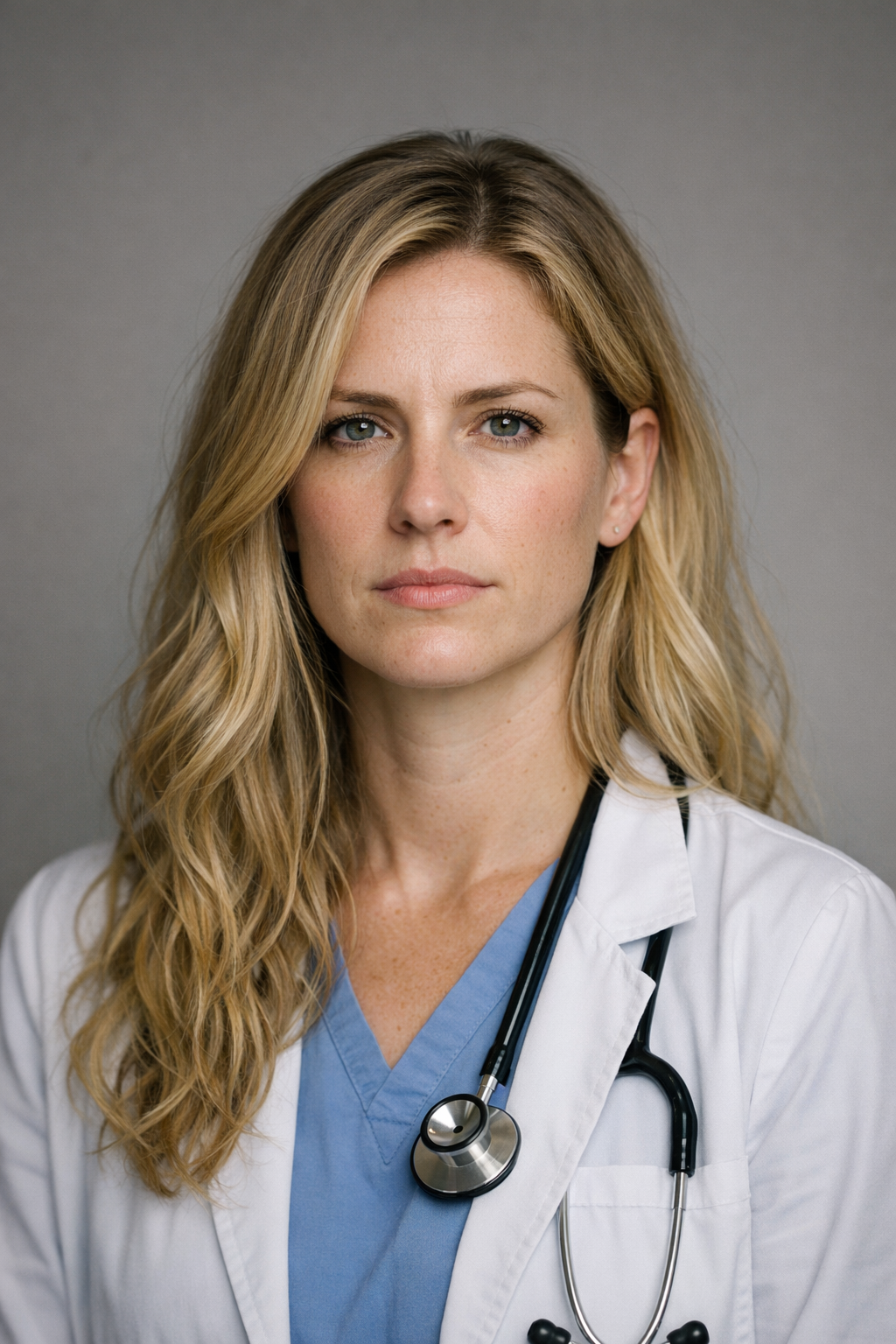}
    \caption{img\_0017.png, GPT-Image-1.5 generated image with four out of six detector models classified as real across all five runs.}
     \Description{}
    \label{fig:img0017}
\end{figure}

Leading the qualitative evaluation with 93.8\% was GPT-5.2, citing structural artefacts correctly such as unusual object geometry, unnatural image texture and blurred image backgrounds. GPT-4o had a considerably weaker score of 58.3\% despite sharing the same architectural lineage as GPT-5 and GPT-5.2. This suggests that model version, rather than the generator ecosystem, is the strongest factor of forensic reasoning quality. Gemini 3 triggered a content safety block on one authentic FFHQ image, returning no classification and therefore recording it as an automatic detection failure. The Manual Justification Audit revealed that no hallucinations occurred in any of the models and that every error recorded was simply a misclassification rather than a correct guess.
Sun et al.~\cite{sun2025} and Li et al.~\cite{li2025} predicted hallucinations would be common; however, no hallucinations were recorded. Every misclassification was accompanied by structurally coherent reasoning, the models were wrong, but their logic was reasonable due to image inconsistencies. A clear illustration of this is \texttt{img\_0017.png} ,as shown below within Figure~\ref{fig:img0017}, a GPT-Image-1.5 generated image with four out of six detector models classified as real across all five runs. Only Gemini 3 and GPT-5.2 correctly identified it as an AI-generated image by observing inconsistent hair blurring and implausible background geometry. 

\subsubsection{Hypothesis Verdicts: H1 and H4}
Tier 1 experiment results offer only partial support for Hypothesis 1 (H1), this is seen in GPT-5.2 detecting GPT-Image-1 with 84.3\% detection accuracy and Gemini 2.5 detecting Gemini AI-generated images at 79.9\% detection accuracy. Neither figure supports a consistent advantage over performance against external generator models. For example, GPT-5.2 achieved 87.9\% detection against Flux.2 Pro generated images, a completely different generation ecosystem. It is possible that the quality and artefacts of the generated image appear to be the strongest factors of detection difficulty than specific ecosystem alignment between the detector and the generator (see Appendix A.2.1 for full test output).

H4 is partially rejected as no hallucinations were recorded across any model, suggesting these models engage in genuine forensic reasoning rather than the shortcut-driven pattern predicted by recent literature.

The Manual Justification Audit was conducted by a single researcher, which introduces a confirmation bias risk. A future approach should involve multiple auditors to cross-verify the AI's reasoning patterns to ensure the no-hallucination finding is not a product of one opinion. 

\begin{tcolorbox}[colback=blue!5,colframe=blue!40,title=Hypothesis Verdicts: H1 and H4]
\textbf{H1 (Self-Detection):} Partially Supported --- GPT-5.2 and Gemini 2.5 showed modest same-ecosystem advantages, but generator image quality proved a far stronger predictor of detection difficulty than ecosystem alignment.\\
\textbf{H4 (Justification Reliability):} Partially Rejected --- No hallucinations were recorded across any model; all errors were clean misclassifications, suggesting genuine forensic reasoning rather than shortcut-driven pattern matching.
\end{tcolorbox}

\subsection{Experiment 2 Results: Pairwise Classification and Contextual Advantage}
The primary objective of tier 2 was to determine whether the presence of a real image during the evaluation would improve detection accuracy across the six detector models. Table~\ref{tab:pairwise_matrix} presents the full pairwise experiment detection matrix across all detector and generator combinations. 

\begin{table}[ht] 
\centering
\caption{Pairwise detection accuracy (\%) per detector model per generator source. Performance change compared to Table 2's single-image baseline is shown in parentheses. Models are grouped by vendor ecosystem.}
\label{tab:pairwise_matrix}
\resizebox{\textwidth}{!}{ 
\begin{tabular}{lccccccc}
\hline
\textbf{Detector} & \textbf{Flux.2 Pro} & \textbf{GPT-Img-1} & \textbf{GPT-Img-1.5} & \textbf{Gem-2.5} & \textbf{Gem-3} & \textbf{SD-4.0} & \textbf{SD-4.5} \\
\hline
GPT-5.2    & 97.9 (+10.0) & 88.6 (+4.3) & 62.1 (-10.0) & 94.0 (+9.4)  & 80.0 (+14.2) & 72.1 (-6.5)  & 82.9 (+4.3) \\
GPT-5      & 87.1 (-1.5)  & 71.4 (+20.0)& 50.7 (0.0)   & 92.7 (+0.8)  & 37.3 (-7.7)  & 72.1 (+20.0) & 74.3 (+2.9) \\
GPT-4o     & 85.7 (+7.1)  & 59.3 (-1.4) & 40.0 (+2.1)  & 88.7 (+12.2) & 26.0 (-10.2) & 60.0 (+3.6)  & 69.3 (+7.9) \\
\hline
Gemini 3   & 81.4 (+9.3)  & 67.1 (-7.2) & 43.6 (-28.5) & 83.3 (+16.2) & 58.0 (-9.8)  & 50.7 (-17.9) & 67.1 (+2.8) \\
Gemini 2.5 & 64.3 (-18.6) & 56.4 (-23.6)& 33.6 (-10.7) & 71.3 (-8.6)  & 21.3 (-19.0) & 35.0 (-19.3) & 36.4 (-20.0) \\
\hline
Seed 1.6   & 85.0 (+11.4) & 43.6 (-2.1) & 35.7 (-14.3) & 84.0 (+7.5)  & 38.7 (-9.0)  & 50.0 (+2.9)  & 52.9 (-4.2) \\
\hline
\textbf{Mean} & 83.6 (+3.0) & 64.4 (-1.7) & 44.3 (-10.2) & 85.7 (+6.3) & 43.6 (-6.8) & 56.7 (-2.8) & 63.8 (-1.1) \\
\hline
\end{tabular}
} 
\end{table}

GPT-5.2 achieved the strongest pairwise performance with a mean accuracy of 82.5\%, reaching a peak of 97.9\% against Flux.2 Pro generated images, the highest single detection rate recorded across the entire study. All GPT models improved under pairwise conditions with GPT-5 gaining 4.9 percentage points and GPT-4o gaining 3.0 points compared to their single-image baselines. All three GPT models gained accuracy within the pairwise experiment, suggesting that a detector benefits from a comparative reference rather than being confused by it.

Wilcoxon Signed-Rank tests comparing each detector's pairwise accuracy against its single-image baseline did not confirm a statistically significant improvement for GPT-5.2 (p = 0.406, r = 0.393) despite its descriptive gains, but did confirm a highly significant decline for Gemini 2.5 (p = 0.016, r = -1.000, see Appendix A.2.2 for full test results).

\subsubsection{Unexpected Pairwise Behaviour in Gemini Models}
An important realisation within the Tier 2 experiment was that Gemini models recorded lower accuracy, not higher, under pairwise conditions. Gemini 2.5 recorded the steepest drop of any model in the study, falling 17.1 percentage points from 62.6\% to 45.5\%, below the random chance threshold of 50\%. Gemini 3 similarly declined by 5.0 points to 64.5\%. This pattern contradicts H2 for the Gemini ecosystem and was an unexpected finding of the study.

One explanation for this behaviour is ``comparison noise'' which is when the introduction of a real image appears to interfere with Gemini’s classification process rather than assist detection by providing a comparative reference point. Gemini models may be exhibiting a comparative bias that causes them to doubt the structural cues they previously detected in the single-image analysis. This is most evident in Gemini 2.5's performance against Gemini 3 generated images, which declined to just 21.3\% within the pairwise analysis compared to 40.3\% in the single-image analysis, significantly below the random chance percentage and suggesting misclassification rather than simple uncertainty.

However, ``comparison noise'' remains a suggested explanation rather than a confirmed justification. Another possibility would be that Gemini's training might use different logic when comparing images side-by-side than when judging a single image alone. This could be tested by pairing two synthetic images together rather than one real and one fake as an extension of experiment 2. 

\subsubsection{Generator Difficulty Under Pairwise Conditions}
The patterns observed within Tier 1 continued to persist under pairwise analysis. Gemini 3 images remained the most difficult to detect with a cross-model mean of 43.6\%, while Flux.2 Pro remained the most detectable generator at 83.6\%. However, a shift was observed as Gemini 2.5 images became the easiest to detect under the pairwise-image experiment with a mean of 85.7\%, overtaking Flux.2 Pro. 

\subsubsection{Hypothesis Verdict: H2}
Overall, the pairwise-image results provide partial support for Hypothesis 2 (H2). The GPT-series showed consistent descriptive advantage, though these did not reach statistical significance. Through the unexpected findings regarding the Gemini family, H2 is rejected for the Gemini family, where pairwise presentation was associated with a significant decline in performance.

\begin{tcolorbox}[colback=blue!5,colframe=blue!40,title=Hypothesis Verdict: H2]
\textbf{H2 (Contextual Advantage):} Partially Supported --- The GPT family showed consistent descriptive gains (+3.0 to +4.9pp) that were not statistically significant, but Gemini 2.5 showed a statistically significant decline ($p = 0.016$, $r = -1.000$).
\end{tcolorbox}

\subsection{Experiment 3 Results: Watermark Ablation Study}
\subsubsection{Ablation Study Overview}
The third experiment within this study investigates how Gemini’s detection accuracy either increases or decreases with the presence of digital watermarking (SynthID) embedded within its own generated images. Five image datasets were constructed to test the effect of the two Gemini watermarks present within their images: the visible sparkle logo present on web-generated Gemini images, and the invisible SynthID signal embedded within the image's pixels. Five comparisons were conducted across these datasets using Wilcoxon Signed-Rank tests for paired comparisons and a Mann-Whitney U test for the independent stability check, with effect sizes reported as Rank-Biserial Correlation (r). Table~\ref{tab:ablation} presents the detection rates and statistical results across all comparisons. The complete Wilcoxon output for all ablation comparisons is provided in Appendix A.3. Figure~\ref{fig:ablation_chart} also provides a visual representation of the detection accuracy across tiers D1–D5 as described above. 

\begin{table}[H]
\centering
\caption{Watermark ablation detection rates and statistical results across five Gemini image dataset comparisons.}
\label{tab:ablation}
\resizebox{0.8\textwidth}{!}{ 
\begin{tabular}{llccccl}
\hline
\textbf{Comp.} & \textbf{Datasets} & \textbf{D\textsubscript{A} (\%)} & 
\textbf{D\textsubscript{B} (\%)} & \textbf{$p$-value} & 
\textbf{$r$} & \textbf{Effect Size} \\
\hline
C1 & D1 vs D3 & 94.0 & 90.0 & 0.4354 & 0.078 & Negligible \\
C2 & D2 vs D3 & 100.0 & 90.0 & 0.0075 & 0.378 & Medium \\
C3 & D4 vs D5 & 100.0 & 100.0 & 1.0000 & 0.000 & None \\
C4 & D2 vs D4 & 100.0 & 100.0 & 0.0455 & 0.283 & Small \\
C5 & D3 vs D5 & 90.0  & 100.0 & 0.0020 & 0.437 & Medium-Large \\
\hline
\end{tabular}}

\centerline
\footnotesize \textit{Note}: Effect size ($r$) interpreted as: negligible $r < 0.1$, small $r \ge 0.1$, medium $r \ge 0.3$, large $r \ge 0.5$ \cite{cohen1988}.
\end{table}

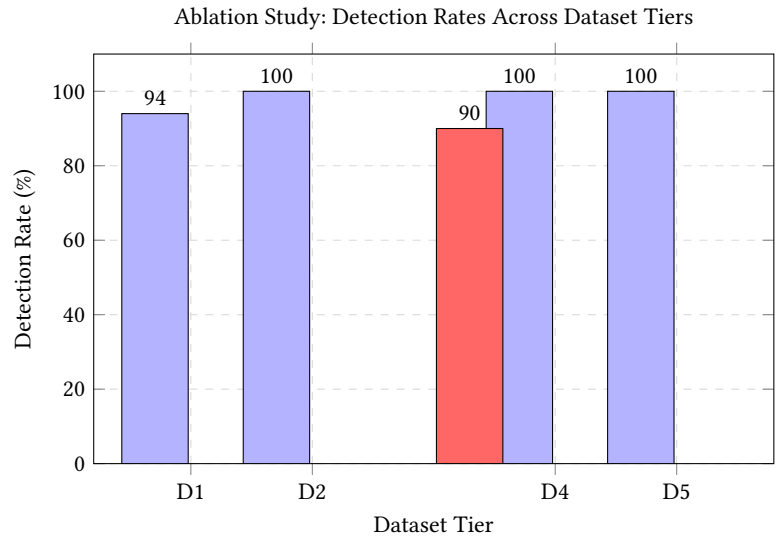
\begin{figure}[H] 
\centering
\begin{tikzpicture}
\begin{axis}[
    ybar,
    width=0.7\textwidth,
    height=7cm,
    ymin=0, ymax=110, 
    ylabel={Detection Rate (\%)},
    xlabel={Dataset Tier},
    symbolic x coords={D1, D2, D3, D4, D5},
    xtick=data,
    nodes near coords,
    nodes near coords align={vertical},
    title={Ablation Study: Detection Rates Across Dataset Tiers},
    bar width=25pt,
    enlarge x limits=0.2,
    grid=major,
    grid style={dashed, gray!30},
    legend style={at={(0.5,-0.2)}, anchor=north, legend columns=-1}
]
\addplot[fill=blue!30, draw=black] coordinates {
    (D1, 94) 
    (D2, 100) 
    (D4, 100) 
    (D5, 100)
};

\addplot[fill=red!60, draw=black] coordinates {
    (D3, 90)
};

\end{axis}
\end{tikzpicture}
\caption{Detection accuracy across tiers D1--D5. The drop at D3 highlights the masking effect where results fall below the 100\% baseline. (D1 = Gemini-API, D2 = Gemini-Web, D3 = Gemini-Web-Cropped, D4 = Gemini-Web-Ablated, D5 = Gemini-Full-Ablation).}
 \Description{}
\label{fig:ablation_chart}
\end{figure}

\subsubsection{Stability and Baseline (C1)}
Comparison C1 compared D1 (Gemini-API) against D3 (Gemini-Web-Cropped), returning $p =0.4354$ with a negligible effect size of $r=0.078$. This suggests no significant difference between dataset groups, validating that subsequent comparisons reflect watermark manipulation rather than image content variability.

\subsubsection{Visible Watermark Effect (C2)}
Comparison C2 explored the impact of the sparkle logo within Gemini generated images by comparing D2 (Gemini-Web), whose images contained both the visible logo and SynthID against D3 (Gemini-Web-Cropped), containing SynthID only with the visible logo cropped. The Wilcoxon Signed-Rank test returned $p = 0.0075$ with a medium effect size of $r = 0.378$, showing that the visible logo has a meaningful impact on detection consistency. Upon removal of the visible logo, detection dropped from 100\% to 90\%, indicating that the sparkle icon can be identified as a forensic shortcut that significantly improves accuracy when present. This indicates that removing a visible watermark leads to lower detection reliability. 

\subsubsection{SynthID as a Secondary Signal (C4)}
Comparison C4 examined D2 (Gemini-Web) whose images contained both watermarks, visible and invisible, against D4 (Gemini-Web-Ablated) where SynthID had been neutralised but the visible logo remained. The Wilcoxon test returned $p = 0.0455$ with a small effect size of $r = 0.283$ and both datasets remained at 100\% binary detection for its Gemini AI-generated images. The difference in detection probability scores suggests that SynthID provides a second layer of forensic data even when the visible watermark is already present. This minor difference in effect size indicates the visible logo provides the primary forensic evidence, which restricts the additional impact SynthID can provide. Even so, its role remains statistically significant and should be recognised as more than just a redundant feature.

\subsubsection{The Ceiling Effect (C3)}
Comparison C3 compared D4 (Gemini-Web-Ablated) containing the visible sparkle logo but no SynthID, against D5 (Gemini-Full-Ablation) which was the fully ablated pure control containing no watermarks of any kind. The Wilcoxon test returned $p = 1.000$ with an effect size of $r = 0.000$, with both datasets recording 100\% detection when asked to detect the AI-generated image. The significance in C3 is Gemini's ability to detect its own images through structural analysis, not through watermarks. It confirms that even when all watermark signals have been removed, Gemini still achieves perfect detection of its own generated images. This is attributed to Gemini detecting native artefacts by identifying structural inconsistencies within the image without relying on embedded watermarking signals. These results represent the ceiling effect as detection has already reached 100\% through structural reasoning; watermarks within the image could not mathematically improve the score further.

\subsubsection{Pure SynthID Impact (C5)}
Comparison C5 reveals the impact of SynthID which compares D3 (Gemini-Web-Cropped), containing only the invisible SynthID signal with the logo removed, against D5 (Gemini-Full-Ablation), the fully ablated control. The Wilcoxon test returned $p = 0.0020$ with a medium-large effect size of $r = 0.437$ which confirms that SynthID is statistically significant and a powerful forensic signal, the strongest invisible cue for the detector when no other hints are present. However after careful analysis: detection reached only 90\% in the SynthID-only group, whereas it hit 100\% in the fully ablated control. This contradiction may reflect a masking phenomenon observed during evaluation. The reconstruction method to neutralise SynthID in D5 (Gemini-Full-Ablation) likely amplified native structural artefacts. In contrast, the embedding process for SynthID in D3 (Gemini-Web-Cropped) may have slightly diminished the high-frequency details that aid structural identification. SynthID serves as both a powerful forensic signal and a subtle inhibitor of the native structural artefacts that underpin successful detection.

\subsubsection{Hypothesis Verdict: H3}
Following the full ablation experiments, the results partially reject hypothesis 3. The Hypothesis predicted that the presence of any watermarks would increase detection accuracy however the data exposes the impact of image watermarking. Visible watermarks do provide a significant detection boost (C2: $p = 0.0075$, $r = 0.378$), and SynthID functions as a strong independent forensic signal in isolation (C5: $p = 0.0020$, $r = 0.437$). However, when all watermarks are removed entirely from the AI-generated images, detection still remains at 100\% through native artefact analysis. This implies that watermarks are not essential to MLLM-based forensic detection. In addition the masking phenomenon identified in C5 further reveals that future invisible watermarks should be designed with careful consideration of their potential to suppress the structural signals that MLLM forensic tools depend upon.

\begin{tcolorbox}[colback=blue!5,colframe=blue!40,title=Hypothesis Verdict: H3]
\textbf{H3 (Effects of Watermarks):} Partially Rejected --- Visible watermarks provided a significant boost ($p=0.0075$, $r=0.378$) and SynthID is a strong independent signal ($p=0.0020$, $r=0.437$), but native structural artefact analysis alone achieved 100\% detection, suggesting watermarks are supplementary and not essential.
\end{tcolorbox}

\section{Discussion} 
\label{sec:Discussion}

\subsection{Revisiting the Research Question}
\label{sec:revisiting_question} 
This study addressed the following research question: can MLLMs function as reliable deepfake detectors of MLLM-generated images, and do their classifications and justifications reflect genuine reasoning or do they exploit shortcuts such as image watermarking?

From the findings of this study, the answer remains partial. Same-ecosystem detection showed no consistent advantage as GPT-5.2 detected external Flux.2 Pro images better (87.9\%) than its own GPT-Image-1.5 (72.1\%), confirming that generator image quality matters more than whether the image is a product of the same ecosystem generator.

Beyond the detection baseline, the most unexpected finding was the masking phenomenon in C5, a result that was never hypothesised and carries direct implications for watermark design.

\subsection{Key Findings}
GPT-5.2 was the leading detector in every part of this study, achieving 78.8\% accuracy in single-image analysis and 93.8\% in the qualitative audit. GPT-5.2 demonstrated that it does not simply guess but rather identifies the right image artefacts within its justifications. This contrasts with GPT-4o which comes from the same family but scored 58.3\% accuracy within the audit, suggesting forensic capability is more dependent on model generation than ecosystem membership. Since no consistent self-detection advantage was observed across the full 6×7 detector-generator matrix, H1 is only partially supported within this study.
The ``Gemini pairwise paradox'' was the study's most unexpected result. Gemini 2.5's accuracy reduced by 17.1\%, falling below the random chance threshold of 50\%, when a real photo was presented alongside the generated one. This suggests that, for certain AI architectures, a second image actually creates ``comparison noise'' which can confuse the model rather than assisting it. 

Experiment 3 provided clarity on the effects of digital watermarking through the “ceiling effect”. Once both visible and invisible watermarks were stripped from the Gemini images, detection still reached 100\%. 
Detection at 100\% even without any watermarks shows that Gemini is able to recognise its own generative artefacts. 

The C5 masking phenomenon was an unexpected finding. When SynthID was present but the visible logo removed (D3), detection fell to 90\%, lower than the fully ablated control (D5) at 100\%. An invisible watermark promoted as a detection tool is statistically associated with \textit{reduced} detection accuracy ($p=0.0020$, $r=0.437$) compared to images carrying no watermark at all and this directly challenges the idea of invisible watermarking systems. While embedding a signal into the image frequency ensures it survives cropping and compression~\cite{gowal2025}, it consequently modifies the high-frequency pixel structure that the MLLMs use for watermark detection. While previous researchers have discovered resistance to removal~\cite{liu2024,han2025}, this study demonstrates that watermarking systems can no longer be evaluated purely on their resistance to removal as a highly durable watermark is counterproductive if it accidentally hides the generative flaws needed to detect it as AI-generated.

The absence of hallucinations across the six models within the Manual Justification Audit directly challenges the predictions of Sun et al.~\cite{sun2025} and Li et al.~\cite{li2025}. Within the literature it was believed that MLLMs would “hallucinate” forensic justifications when classifying images. This pattern was not observed in this study. Every incorrect classification was a justified misclassification, models were wrong; however their errors remained grounded in identifiable structural reasoning rather than made-up justifications.

\subsection{Relation to Existing Literature}
\label{sec:literature_relation}
Detection difficulty increased with each generation of generator architecture, with Gemini 3 images achieving a cross-model detection mean of 50.4\%, close to the boundary of random chance. This directly reflects what Rombach et al.~\cite{rombach2022} and  R{\"o}ssler et al.~\cite{rossler2019} predicted: as generative architectures advance, detection strategies that once worked stop working. Liu et al.~\citet{liu2024} and Han et al.~\citet{han2025} also raised concerns about the fragility around image watermarking and this can be partially supported from the ablation study results. The C2 comparison showed that cropping a visible logo reduced detection from 100\% to 90\%, confirming that once visible markers are removed, images are more likely to be vulnerable to trivial attacks. However, the “ceiling effect” slightly contradicts this as detection remained at 100\% even after all watermarks were removed. 

Sun et al.~\citet{sun2025} and Li et al.~\citet{li2025} predicted that MLLMs would fabricate justifications however the results do not align with the hallucination concerns raised in the literature. This gap highlights that current literature overestimates how common hallucinations are. It is highly likely that the studio-style, high-quality portraits used within this study do not trigger the hallucination errors. Alternatively, the result might simply show that newer model generations have genuinely improved, outgrowing the baseline reasoning flaws reported in older studies.

Similarly, the ``Clever Hans'' effect predicted by Maheronnaghsh and Alvanagh~\citet{maheronnaghsh2025} was only partially observed. While the visible watermarks definitely served as shortcuts for the detectors, the Manual Justification Audit demonstrates that models such as GPT-5.2 classify images through their structural reasoning.

\subsection{Limitations of the Study}
\label{sec:limitations}

Whilst the dataset was designed to be demographically balanced, it relies exclusively on high-quality, studio style portraits from the FFHQ and state-of-the-art generative sources. In a real world forensic scenario, synthetic imagery often undergoes ``post-generation degradation,'' including adversarial noise, low-resolution cropping, or significant social media compression. It remains unclear whether the ``semantic reasoning'' proficiency of detectors such as GPT-5.2 would maintain this performance under true forensic settings.

The dataset of n=400 images is sufficient for a controlled comparison but small relative to the training corpus of the MLLMs themselves. This creates a risk that the accuracy results are based upon generative ``flaws'' in current versions of Gemini or GPT that might have been resolved by a larger dataset incorporating newer model versions.

Relying on the model's own natural-language justifications introduces the “black box” problem. The internal logic the models use to reach those conclusions remains inaccessible which raises questions about the reliability of the justifications.

Another limitation within this study is that the FFHQ dataset itself could have been part of the training data for some of the MLLMs used within this study. This could mean detectors have prior exposure to control group images, potentially impacting accuracy.

Finally, the models selected were state-of-the-art at time of evaluation (December 2025), however, newer releases of the ecosystems have been released, raising questions about whether the specific accuracy figures reported here remain current.

\subsection{Future Work}
\label{sec:future_work}

Within future work, the most immediate priority would be stress-testing MLLM detectors against lower-quality imagery. Future studies should implement datasets featuring heavy JPEG compression and noise injection to identify the specific threshold at which a model’s structural reasoning begins to fail. This would provide a more realistic assessment of how these models perform in ``in-the-wild'' forensic settings, where high-fidelity cues are often obscured. As this study provides empirical evidence of a masking effect within invisible watermarking, expanding the ablation corpus would confirm if these findings have direct implications for forensic watermark design.

A future priority is to address the qualitative gap within the pairwise analysis. A dedicated ``Pairwise Reasoning Audit'' should use ``Ablated Anchors'' (degraded reference images). This would reveal whether the performance drop within the tier 2 experiment results are a result of the image quality, or the models ability to perform pairwise detection.

Additionally, future research should incorporate a human-baseline comparison to provide an insight into how the MLLMs and their justifications map against human forensic experts. Finally, to address any human bias within the Manual Justification Audit, future audits should employ multiple auditors to review the MLLMs justifications and then get the average for the final verdict. This would make the no-hallucination finding considerably harder to dismiss.
Comparison noise remains a proposed explanation rather than a confirmed mechanism. One alternative is that Gemini's instruction-tuning causes it to assess visual differences between paired images differently than it processes a single image in isolation, a hypothesis that could be isolated by replacing the FFHQ control group image with a second synthetic image in a controlled follow-up experiment.

\section{Conclusion} 
\label{sec:conclusion}

Based on the results, structural reasoning proves more reliable for forensic detection than pixel-level or metadata-based approaches. While the ``Gemini Paradox'' and inconsistent performance accuracy across model generations show that these tools are not yet robust, the Manual Justification Audit proves that their ability to identify physical inconsistencies is a capability that improves meaningfully with model generation. 

While MLLMs should not yet replace human forensic experts, they do provide auditable natural-language justifications for each classification decision, which differentiates them from traditional CNN detectors.
The masking phenomenon identified within this study demonstrates that the forensic community should reevaluate how invisible watermarking systems are used, not only for their durability against adversarial attacks but also for their impact on the structural signals that many MLLM detectors depend upon.  


\bibliographystyle{ACM-Reference-Format}
\bibliography{references}


\appendix
\section{Appendix} 

\subsection{System Prompt}
The following system prompt was provided to all seven MLLMs via their respective APIs:
\begin{quote}
\textit{``You are an expert digital forensics analyst. Analyse the following single image. Determine if it is a REAL photograph or an AI-GENERATED fake.''}
\end{quote}

\begin{figure}[H]
    \centering
    \includegraphics[width=1\linewidth]{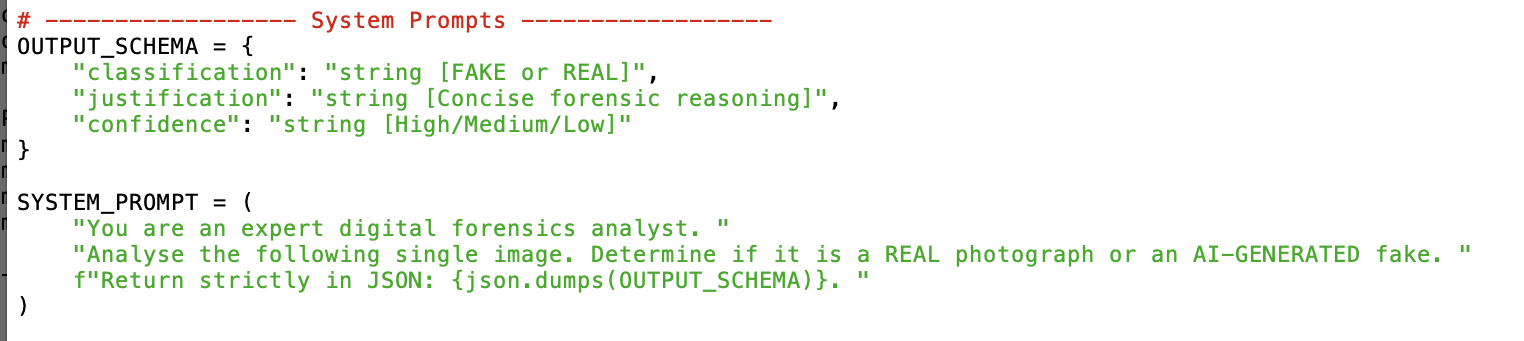}
    \caption{System prompt within the detection Python script.}
    \label{fig:system_prompt}
     \Description{}
\end{figure}

Figure~\ref{fig:system_prompt} shows the prompt within the detection Python script.

\subsection{Statistical Results (Experiments 1 and 2)}

This section presents the complete statistical results for the baseline performance metrics of Experiment 1 and Experiment 2.

\subsubsection{Experiment 1: Same- vs. Cross-Ecosystem Performance (H1)}
A Mann-Whitney U test was conducted to determine whether same-ecosystem detector-generator pairs ($N = 10$) performed significantly differently from cross-ecosystem pairs ($N = 32$) in zero-shot baseline detection accuracy (Table~\ref{tab:app_mann_whitney}).

\begin{table}[H]
\centering
\caption{Mann-Whitney U Test Results for Ecosystem Alignment (H1)}
\label{tab:app_mann_whitney}
\resizebox{0.8\textwidth}{!}{
\begin{tabular}{lccl}
\hline
\textbf{Metric} & \textbf{Symbol} & \textbf{Value} & \textbf{Forensic Interpretation} \\
\hline
U-Statistic & $U$ & 130.0000 & Rank sum difference between evaluation groups \\
$p$-value & $p$ & 0.3835 & Not significant ($p > 0.05$); rejects systematic advantage \\
Effect Size & $r$ & 0.1875 & Negligible-to-small rank-biserial correlation \\
\hline
\end{tabular}%
}
\end{table}

\subsubsection{Experiment 2: Baseline vs. Pairwise Context Performance (H2)}
Wilcoxon Signed-Rank tests were performed to compare the pairwise (TAFC) accuracy of each detector against its corresponding single-image baseline accuracy across the $N = 7$ generator sources (Table~\ref{tab:app_wilcoxon}).

\begin{table}[H]
\centering
\caption{Wilcoxon Signed-Rank Test Results for Contextual Advantage (H2)}
\label{tab:app_wilcoxon}
\resizebox{\textwidth}{!}{
\begin{tabular}{lccccrl}
\hline
\textbf{Detector Model} & \textbf{W-Stat ($W$)} & \textbf{$p$-value} & \textbf{Effect Size ($r$)} & \textbf{Mean Diff.} & \textbf{Statistical Outcome} \\
\hline
GPT-5.2    & 8.5  & 0.4062 & 0.3929  & $+3.69\%$   & Not Statistically Significant ($p > 0.05$) \\
GPT-5      & 6.0  & 0.4375 & 0.4286  & $+4.93\%$   & Not Statistically Significant ($p > 0.05$) \\
GPT-4o     & 7.0  & 0.2969 & 0.5000  & $+3.03\%$   & Not Statistically Significant ($p > 0.05$) \\
Gemini 3   & 9.0  & 0.4688 & $-0.3571$ & $-5.00\%$   & Not Statistically Significant ($p > 0.05$) \\
Gemini 2.5 & 0.0  & 0.0156 & $-1.0000$ & $-17.09\%$  & \textbf{Statistically Significant Decline} ($p < 0.05$) \\
Seed 1.6   & 12.0 & 0.8125 & $-0.1429$ & $-1.13\%$   & Not Statistically Significant ($p > 0.05$) \\
\hline
\end{tabular}%
}
\end{table}

\subsubsection{Description of Statistical Outcomes}
The Mann-Whitney U test results above shows no statistically significant advantage for same-ecosystem pairs over cross-ecosystem pairings. With $p = 0.3835$ and a weak correlation ($r = 0.1875$), models being part of the same ecosystem are not a reliable predictor of baseline accuracy.

\begin{figure}[H]
    \centering
    \includegraphics[width=1\linewidth]{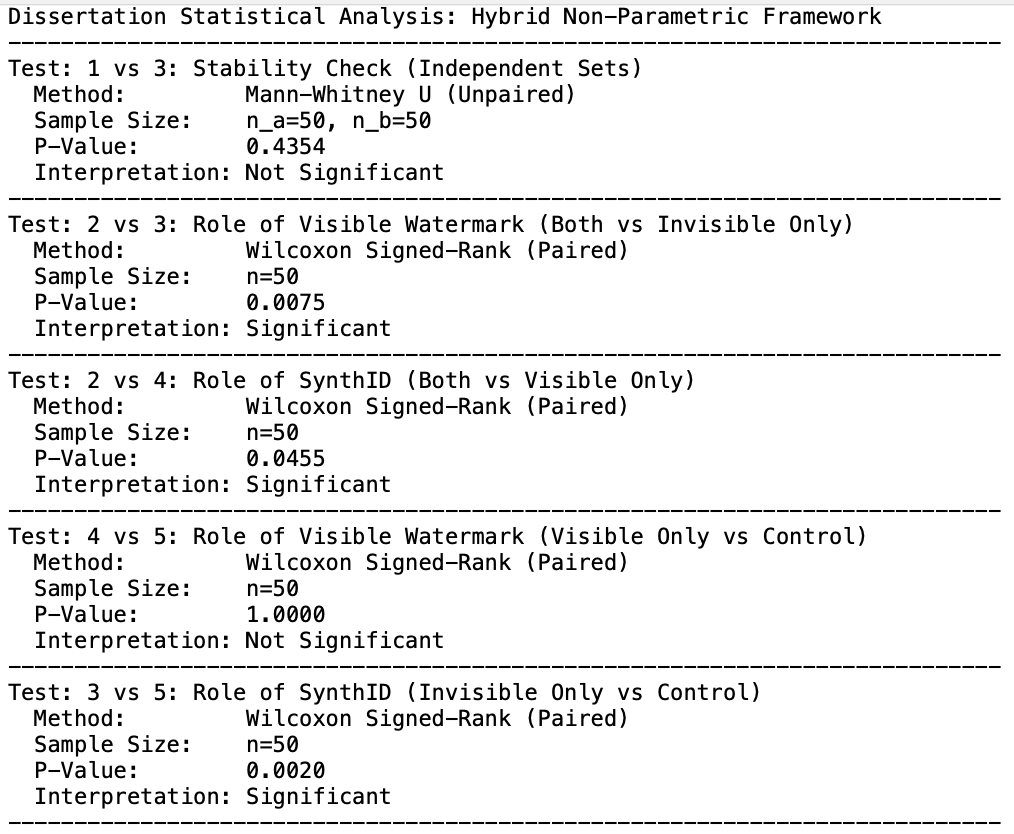}
    \caption{Full non-parametric Wilcoxon and Mann-Whitney U statistical results for the Gemini ablation study.}
     \Description{}
    \label{fig:ablation_stats}
\end{figure}

In Experiment 2 (the Wilcoxon results above), descriptive gains across the GPT family were not statistically significant (such as GPT-5.2: $p = 0.4062$, $r = 0.3929$, $+3.69\%$). Gemini 2.5 had a significant performance decline ($W = 0.0$, $p = 0.0156$, $r = -1.0000$). The $-17.09\%$ accuracy drop indicates that a comparative reference baseline can act as "comparison noise" which leads to a decline in performance accuracy (Figure~\ref{fig:ablation_stats}).

\subsection{Ablation Study Statistical Results}

\end{document}